\pdfoutput=1
\newif\ifaclstyle
\aclstylefalse
\newif\ifanonymous
\anonymousfalse %
\ifaclstyle
  \documentclass[11pt]{article}
  \usepackage[review]{acl}
  \usepackage{times}
  \anonymoustrue
\else
  \documentclass[10pt]{article}
  \usepackage[letterpaper,top=0.70in,bottom=0.78in,left=0.68in,right=0.68in]{geometry}
\fi
\usepackage[T1]{fontenc}
\usepackage[utf8]{inputenc}
\ifaclstyle\else\usepackage{lmodern}\fi
\usepackage{microtype}
\usepackage{amsmath,amssymb,mathtools}
\usepackage{booktabs,longtable}
\usepackage{balance}
\usepackage{afterpage}
\usepackage{etoolbox}
\makeatletter
\patchcmd{\@BAlancecol}{\setbox\@leftcolumn}
  {\advance\@tempdima by 2pt\setbox\@leftcolumn}{}{}
\makeatother
\usepackage{multirow}
\usepackage{array}
\newcolumntype{P}[1]{>{\RaggedRight\arraybackslash}p{#1}}
\newcolumntype{Y}{>{\RaggedRight\arraybackslash}X}
\usepackage{graphicx}
\usepackage{xcolor}
\usepackage{url}
\usepackage{hyperref}
\usepackage{enumitem}
\usepackage{caption}
\usepackage{subcaption}
\usepackage{siunitx}
\usepackage{natbib}
\usepackage{placeins}
\usepackage{float}
\usepackage{needspace}
\usepackage{ragged2e,multicol}
\ifaclstyle\else
\usepackage{titlesec}
\titleformat{\section}{\normalfont\fontsize{12}{14}\selectfont\bfseries\raggedright}{\thesection}{.65em}{}
\titleformat{\subsection}{\normalfont\fontsize{10}{12}\selectfont\bfseries\raggedright}{\thesubsection}{.65em}{}
\titlespacing*{\section}{0pt}{9pt plus 2pt minus 2pt}{4pt plus 1pt}
\titlespacing*{\subsection}{0pt}{7pt plus 2pt minus 1pt}{3pt plus 1pt}
\titlespacing*{\paragraph}{0pt}{5pt plus 1pt minus 1pt}{.65em}
\fi

\usepackage{tikz}
\usetikzlibrary{arrows.meta,positioning,fit,calc}
\usepackage{pgfplots}
\usepgfplotslibrary{groupplots}
\pgfplotsset{compat=1.14,
  every axis/.append style={font=\rmfamily\fontsize{8}{9.5}\selectfont},
  every axis/.append style={
    /pgf/number format/1000 sep={},
    every node near coord/.append style={/pgf/number format/fixed,
      /pgf/number format/precision=2},
    legend cell align=left,
    legend image post style={line width=.65pt}
  }}
\usepackage{tabularx}
\IfFileExists{xurl.sty}{\usepackage{xurl}}{%
  \appto\UrlBreaks{\do\a\do\b\do\c\do\d\do\e\do\f\do\g\do\h\do\i\do\j\do\k\do\l\do\m\do\n\do\o\do\p\do\q\do\r\do\s\do\t\do\u\do\v\do\w\do\x\do\y\do\z\do\0\do\1\do\2\do\3\do\4\do\5\do\6\do\7\do\8\do\9}
}

\hypersetup{
  pdftitle={Seen but Unsaid: Locating and Repairing Cross-Image Comparison in Vision--Language Models},
  pdfsubject={Cross-image correspondence in vision--language models},
  bookmarksnumbered=true,
  pdfdisplaydoctitle=true,
  colorlinks=true,
  linkcolor=sbuLink,
  citecolor=sbuLink,
  urlcolor=sbuLink
}
\ifanonymous\else\hypersetup{pdfauthor={Zhaoning Zhang}}\fi

\ifaclstyle\else
\fi
\ifaclstyle\else
\setlist[itemize]{leftmargin=*,topsep=2pt,itemsep=1pt,parsep=0pt}
\setlist[enumerate]{leftmargin=*,topsep=2pt,itemsep=1pt,parsep=0pt}
\fi
\ifaclstyle
  \DeclareCaptionFont{sbucaption}{\fontsize{10}{12}\selectfont}
\else
  \DeclareCaptionFont{sbucaption}{\fontsize{9}{10.8}\selectfont}
\fi
\ifdefined\XeTeXversion
  \microtypesetup{protrusion=true,expansion=false}
\else
  \microtypesetup{protrusion=true,expansion=true}
\fi
\ifaclstyle\else
\fi
\ifaclstyle\else
\fi

\newcommand{\pp}{\,pp}
\newcommand{\qwen}[1]{\textsc{Qwen3-VL-#1}}
\newcommand{\ivl}{\textsc{InternVL3.5-8B}}
\newcommand{\gemma}{\textsc{Gemma-3-12B}}
\newcommand{\syn}{\textsc{Syn}}
\newcommand{\hp}{\textsc{HP}}
\newcommand{\spt}{\textsc{SP}}
\newcommand{\refm}{\textsc{ref}}
\newcommand{\dref}{\Delta_{\textsc{ref}}}
\newcommand{\rref}{\rho_{\textsc{ref}}}
\newcommand{\Acc}{\mathrm{Acc}}
\newcommand{\Raff}{\mathrm{Acc}_{\mathrm{cal}}}
\newcommand{\Pans}{\mathrm{Probe}_{\mathrm{ans}}}
\newcommand{\Preg}{\mathrm{Probe}_{\mathrm{reg}}}
\newcommand{\MaxSim}{\mathrm{MaxSim}}
\newcommand{\Ppair}{\mathrm{Probe}_{\mathrm{pair}}}

\ifaclstyle
\title{Seen but Unsaid: Locating and Repairing\\
Cross-Image Comparison in Vision--Language Models}
\else
\title{
\fontsize{15}{17.5}\selectfont
\textbf{
Seen but Unsaid: Locating and Repairing\\
Cross-Image Comparison in Vision--Language Models
}
}
\fi

\ifanonymous
  \author{Anonymous Author(s)}
\else
  \author{Zhaoning Zhang\\
  University of Toronto\\
  \texttt{zhaoning.zhang@mail.utoronto.ca}}
\fi

\date{}


\ifaclstyle\else
\makeatletter
\patchcmd{\@maketitle}{\vskip 2em}{\vskip .6em}{}{}
\patchcmd{\@maketitle}{\vskip 1.5em}{\vskip .8em}{}{}
\patchcmd{\@maketitle}{\vskip 1em}{\vskip .4em}{}{}
\patchcmd{\@maketitle}{\vskip 1.5em}{\vskip .6em}{}{}
\makeatother
\fi

\definecolor{sbuLink}{HTML}{2B4C6F}
\definecolor{sbuTeal}{HTML}{1F6F78}
\definecolor{sbuBrick}{HTML}{B5523B}
\definecolor{sbuSlate}{HTML}{5B6670}
\definecolor{sbuSand}{HTML}{F3EEE4}
\definecolor{sbuLine}{HTML}{8A949C}
\pgfplotsset{every axis/.append style={
  scaled ticks=false,
  tick label style={font=\fontsize{7.5}{9}\selectfont},
  label style={font=\fontsize{8}{9.5}\selectfont},
  legend style={font=\fontsize{7.5}{9}\selectfont},
  legend cell align=left,
  every node near coord/.append style={/pgf/number format/fixed,
    /pgf/number format/precision=2},
}}

\begin{document}
\twocolumn
\maketitle
\begin{abstract}
Vision--language models (VLMs) often fail to match a point or an object across two images, even when the match can be decoded from their hidden states. We trace where this information is lost in \qwen{2B/4B/8B/32B}, \ivl{}, and \gemma{} on synthetic shape matching, HPatches point correspondence, and SPair-71k semantic correspondence, comparing input-matched probes with calibrated outputs. The models fail at two distinct points. Exact copies are compared and the result reaches the answer position, but the output does not use it, whether or not the shapes have names. Semantic correspondences are present in the visual tokens but are only weakly compared and never reach the answer position. The comparison runs through 16 to 32 attention heads that let tokens of the second image attend to the first. Redistributing their attention at fixed mass removes up to 32 points of accuracy, against at most 7 for random or attention-matched heads, and their attention is gated by salience and weighted by content. Training only these heads (0.02\% of the parameters) recovers 25 to 28 points on SPair-71k and transfers to HPatches and BLINK. Unconstrained LoRA adapters, which reach higher accuracy, converge on the same heads: inserting only these heads' attention into the base model recovers 21--51\% of an adapter's gain, mostly through changed inputs to the heads. Across all 14 trained arms, the selectivity of this small head set tracks the accuracy they reach, and it is read directly from attention, without probes or retraining. The gain in selectivity is not a by-product of more attention to the first image: it comes mostly from the wrong options no longer being drawn to the marked reference.
\end{abstract}

\section{Introduction}
\label{sec:intro}

\begin{figure*}[!t]
\centering
\begin{tikzpicture}[
  font=\rmfamily\fontsize{8}{9.6}\selectfont,
  frame/.style={draw=sbuLine,line width=.5pt,fill=sbuSand!45},
  ttl/.style={anchor=south west,inner sep=1pt,font=\rmfamily\bfseries\fontsize{8}{9.6}\selectfont,text=sbuSlate},
  pt/.style={circle,inner sep=0pt,minimum size=5pt},
  stage/.style={draw=sbuLine,line width=.5pt,align=center,inner sep=3pt,minimum height=.9cm,text width=2.55cm},
  hdr/.style={align=center,font=\rmfamily\bfseries\fontsize{7.6}{9}\selectfont,text=sbuSlate},
  rowlab/.style={anchor=west,align=left,font=\rmfamily\fontsize{7.8}{9.2}\selectfont},
  cell/.style={rounded corners=1pt,minimum width=2.6cm,minimum height=.52cm,inner sep=0pt,font=\rmfamily\fontsize{7.8}{9}\selectfont},
  strong/.style={cell,fill=sbuTeal!78,text=white},
  partial/.style={cell,fill=sbuTeal!28,text=sbuSlate!40!black},
  weak/.style={cell,fill=white,draw=sbuBrick,line width=.6pt,text=sbuBrick!80!black},
  brk/.style={draw=sbuBrick,line width=1.3pt},
  >={Stealth[length=1.6mm,width=1.2mm]}
]
\draw[frame] (0,0) rectangle (4.3,2.5);

\node[pt,fill=sbuBrick] (ref) at (1.25,1.45) {};
\node[anchor=west,inner sep=1pt,font=\rmfamily\scshape\fontsize{7.5}{9}\selectfont,text=sbuBrick] at (1.36,1.45) {ref};
\foreach \p in {(3.0,1.95),(3.35,0.7),(0.85,0.55)} \node[pt,fill=sbuLine!70] at \p {};
\node[anchor=north,font=\rmfamily\fontsize{7.6}{9}\selectfont,text=sbuSlate] at (2.15,-0.05) {\textbf{Image 1} (reference), \emph{encoded separately}};

\draw[frame] (6.6,0) rectangle (10.9,2.5);

\node[pt,fill=sbuSlate] (A) at (7.15,1.95) {}; \node[anchor=west,inner sep=1.5pt] at (A.east) {A};
\node[pt,fill=sbuTeal]  (B) at (9.15,1.75) {}; \node[anchor=west,inner sep=1.5pt,text=sbuTeal] at (B.east) {\textbf{B}};
\node[pt,fill=sbuSlate] (C) at (7.75,0.6)  {}; \node[anchor=west,inner sep=1.5pt] at (C.east) {C};
\node[pt,fill=sbuSlate] (D) at (10.1,0.75) {}; \node[anchor=west,inner sep=1.5pt] at (D.east) {D};
\node[anchor=north,font=\rmfamily\fontsize{7.6}{9}\selectfont,text=sbuSlate] at (8.75,-0.05) {\textbf{Image 2} (candidates), \emph{encoded separately}};

\draw[->,sbuTeal,line width=1.3pt] (B) .. controls (8.0,3.35) and (2.6,3.2) .. (ref);
\foreach \o in {A,C,D} \draw[->,sbuLine,line width=.5pt,densely dashed] (\o) .. controls ($(\o)+(-1.6,1.1)$) and (3.0,2.2) .. (ref);
\node[align=center,font=\rmfamily\fontsize{7.6}{9}\selectfont,text=sbuTeal!70!black] at (5.45,3.6)
  {cross-image heads: each option attends back to salient candidates\\ in image~1, weighted by content similarity};

\node[stage,fill=white] (ans) at (12.95,1.25) {\textbf{Answer position}\\ $\Pans$};
\node[stage,fill=white] (let) at (16.1,1.25) {\textbf{Answer letter}\\ $\Acc$, $\Raff$};
\draw[->,sbuSlate,line width=.7pt] (10.9,1.25) -- (ans.west);
\draw[->,sbuSlate,line width=.7pt] (ans.east) -- (let.west);

\def\ya{-1.25}
\node[hdr] at (5.6,\ya) {Visual tokens\\[-1pt] {\normalfont $\Preg-\Pans$}};
\node[hdr] at (8.55,\ya) {Comparison\\[-1pt] {\normalfont $\rref$}};
\node[hdr] at (11.5,\ya) {Answer position\\[-1pt] {\normalfont $\Pans$}};
\node[hdr] at (14.45,\ya) {Output\\[-1pt] {\normalfont $\Raff$}};
\draw[sbuLine,line width=.4pt] (0,-1.75)--(16.1+1.1,-1.75);

\node[rowlab] at (0,-2.15) {\textbf{Exact copies} (\syn{})};
\node[strong]  at (5.6,-2.15) {$+36.0\pp$};
\node[strong]  at (8.55,-2.15) {10.36};
\node[partial] at (11.5,-2.15) {.64};
\node[weak]    at (14.45,-2.15) {.43};
\draw[brk] (12.97,-1.92)--(12.97,-2.38);

\node[rowlab] at (0,-2.85) {\textbf{Same instance} (\hp{})};
\node[strong]  at (5.6,-2.85) {$+5.8\pp$};
\node[partial] at (8.55,-2.85) {1.88};
\node[strong]  at (11.5,-2.85) {.80};
\node[strong]  at (14.45,-2.85) {.78};

\node[rowlab] at (0,-3.55) {\textbf{Semantic} (\spt{})};
\node[strong]  at (5.6,-3.55) {$+16.3\pp$};
\node[weak]    at (8.55,-3.55) {1.30};
\node[weak]    at (11.5,-3.55) {.43};
\node[weak]    at (14.45,-3.55) {.46};
\draw[brk] (7.075,-3.32)--(7.075,-3.78);

\node[strong,minimum width=.5cm] at (5.0,-4.3) {}; \node[anchor=west] at (5.3,-4.3) {carried};
\node[partial,minimum width=.5cm] at (6.9,-4.3) {}; \node[anchor=west] at (7.2,-4.3) {partly carried};
\node[weak,minimum width=.5cm] at (9.9,-4.3) {}; \node[anchor=west] at (10.2,-4.3) {weak or lost};
\draw[brk] (12.6,-4.12)--(12.6,-4.48); \node[anchor=west] at (12.7,-4.3) {where the signal breaks};
\end{tikzpicture}
\caption{\textbf{Where correspondence information is lost.} \emph{Top:} the task. Each image is encoded separately; under causal attention, tokens of each option in image~2 attend back to salient candidate locations in image~1, and the true option (here B) attends more strongly to the reference. \emph{Bottom:} four measurements along this path for \qwen{8B} on test items ($\rref$ on held-out items). Exact copies are compared strongly and reach the answer position but are not emitted; semantic correspondences are present in the visual tokens but only weakly compared. Shading summarizes the values shown. The cross-image heads that carry the comparison are causal (\S\ref{sec:res_heads}), and training them repairs semantic correspondence (\S\ref{sec:repair}).}
\label{fig:overview}
\end{figure*}
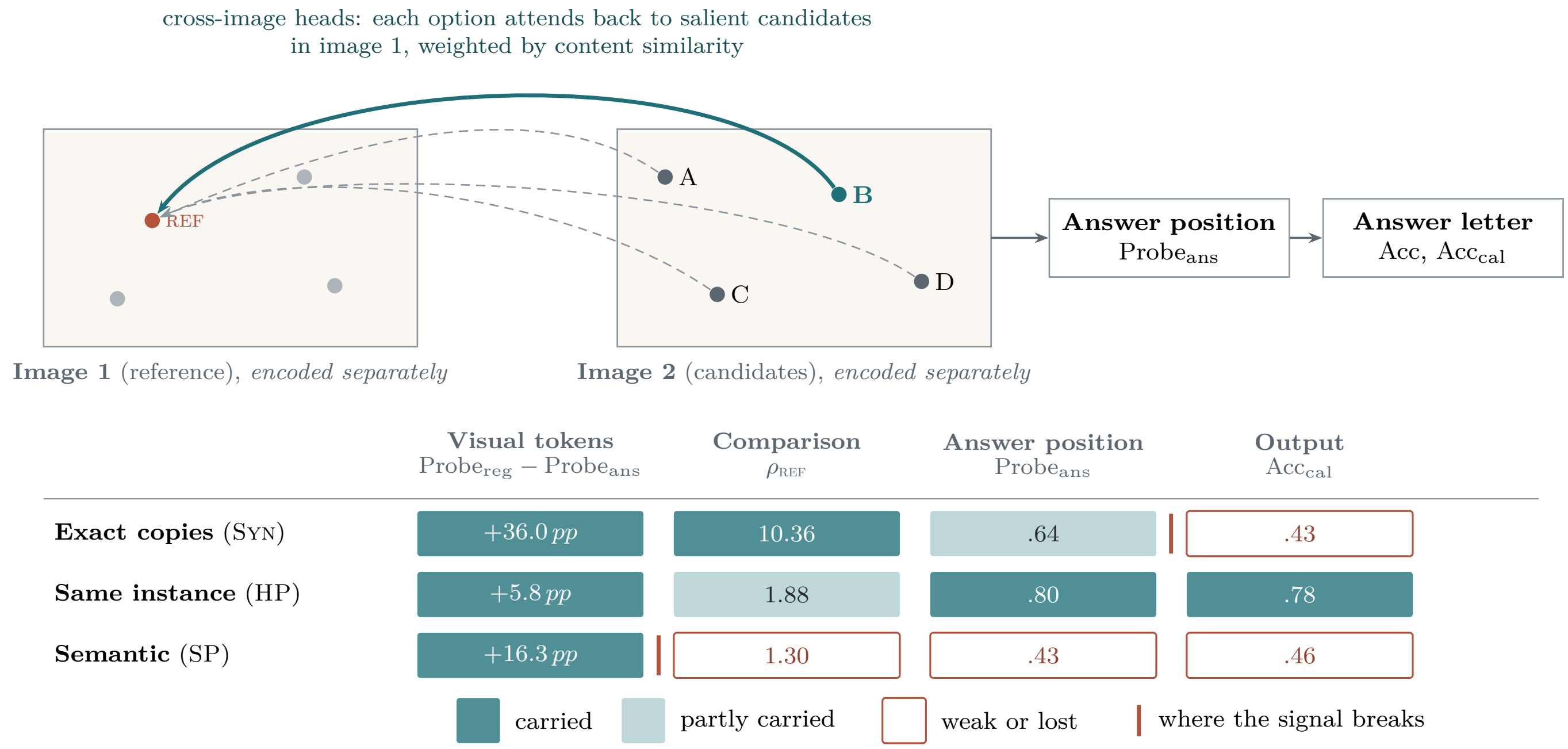

Vision--language models (VLMs) answer open-ended questions about images, yet remain unreliable when a task requires matching a point, a shape, or an object across two images \citep{fu2024blink}. Many such failures do not originate in the vision encoder. VLMs underperform direct readouts of their own encoders on vision-centric tasks \citep{fu2025hidden}; image value tokens inside the language model support semantic correspondence zero-shot even when the model's answer is wrong \citep{liu2025visualrep}; counting fails even when numerical information remains linearly separable \citep{pang2026counting}; and visual evidence can be encoded about as strongly when the answer is wrong as when it is right \citep{nooralahzadeh2026arbitration}. VLMs often \emph{see} more than they \emph{say}.

For cross-image correspondence, \citet{shahgir2026words} argue that this gap is widest for entities without names: on synthetic ``squiggle'' correspondence, \qwen{2B} answers 29.0\% correctly while a representation probe over its hidden states reaches 74.2\%, and teaching arbitrary names narrows the gap. That readout compares hidden states at the annotated locations of the reference and the options, locations the answering model must find for itself. It is a useful upper reference. What it leaves open is the size of the gap when the readout sees exactly what the model sees, the place inside the language model where the information is lost, and the component that would recover it.

Open VLMs encode each image independently, so the comparison between images has to happen inside the language model \citep{wang2026stateful}, where tokens of the second image can retrieve evidence from the first only through attention. The comparison is therefore a computation with a definite location, and we follow the signal along this path (Fig.~\ref{fig:overview}). At each stage we measure what is available with probes that see only what the model sees, and we read every gap against a calibrated output. We then identify the heads that carry the comparison, characterize what they attend to, and test the localization three ways: by training only those heads, by comparing them with matched alternatives, and by asking whether unconstrained fine-tuning converges on the same heads. The main claims are confirmed on held-out items that played no part in any selection.

Locating the loss matters because each locus calls for its own remedy. If the answer is computed but not emitted, the fix must act on how the output reads it; if the information never reaches the answer position, it must act earlier. The same models exhibit both cases, on different tasks.

\noindent Our contributions are as follows.
\begin{itemize}
  \item \textbf{Measurement.} With input-matched probes and a calibrated output baseline, the gap on synthetic matching is the same for nameable and unnameable shapes, and annotated-region readouts on this task measure pixel identity (\S\ref{sec:res_ladder}).
  \item \textbf{Two failure points.} Exact copies are compared, but the result is not read out; semantic correspondences are present in the visual tokens but do not reach the answer position. Both patterns hold in all three families (\S\ref{sec:res_breadth}--\ref{sec:res_priv}).
  \item \textbf{Mechanism.} A small set of cross-image heads carries the comparison. Redistributing their cross-image attention at fixed total mass removes $15$--$32\pp$ of synthetic and same-instance accuracy, against at most $7\pp$ for random or attention-matched heads. Their attention is gated by salience and weighted by content, and on held-out items its selectivity predicts whether the comparison reaches the answer position (\S\ref{sec:heads}).
  \item \textbf{Repair and convergence.} Training only the query and key projections of these heads recovers most of the semantic-correspondence deficit ($+25$ to $+28\pp$), outperforms layer- and attention-matched heads, and transfers to other benchmarks. Unconstrained LoRA adapters, which reach higher accuracy, converge on the same heads: attention transplantation shows that they reshape these heads' attention, mainly through their inputs. Across all 14 trained arms, head selectivity tracks accuracy, giving a probe-free readout of what fine-tuning changes in the comparison step; the gain holds under a scale-free ratio and comes mostly from distractors no longer attending to the marked reference (\S\ref{sec:repair}).
\end{itemize}

\section{Related Work}
\label{sec:related}

\paragraph{Representation versus utilization.}
Benchmarks expose VLM failures on perception tasks that are easy for humans, including visual and semantic correspondence \citep{fu2024blink}. \citet{fu2025hidden} show that encoder readouts outperform the VLM on vision-centric tasks and attribute the bottleneck to the language model's use of visual information. \citet{liu2025visualrep} find that image value tokens inside the language model support segmentation and semantic correspondence zero-shot and that correct perceptual information is often present but not surfaced. \citet{pang2026counting} place the counting bottleneck in the mapping to number tokens, and \citet{veikherman2026encoded} bound how often errors are encoded but unused. \citet{shahgir2026words} link the gap to nameability, and SOCO reports a similar asymmetry between visual and textual queries on real images \citep{dunkel2026soco}; \citet{qiu2026invariance} find that VLM judgments of visual sameness rely on semantic labels. We measure the gap with input-matched probes, show that it is not specific to unnameable entities, and identify the computation step at which it arises.

\paragraph{Readouts, probes, and calibration.}
Probe accuracy must be read against control tasks \citep{hewitt2019control}, and decodability differences across layers do not add up to information lost \citep{xu2020vinfo}. In text-only models, hidden-state probes can recover answers when sequence scoring collapses because of output biases \citep{yan2026wrong}, so a gap that output calibration closes is not evidence of a deeper failure. In VLMs, probe decodability and steering recovery can overstate grounding unless the image is ablated \citep{liao2026decodable}. We report every gap against a calibrated output and include image-ablation controls.

\paragraph{Attention heads in VLMs and multiple-choice answering.}
Activation patching and mediation analysis \citep{vig2020causal,meng2022rome} have located where visual evidence gains or loses influence in single-image settings: arbitration between percept and prior \citep{nooralahzadeh2026arbitration}, heads through which priors override vision \citep{lietzow2026vision}, object localization \citep{schaumloffel2026localization}, and visual entities resolved too late for factual recall \citep{venhoff2025toolate,biran2024hopping}. Specialized heads ground text to image regions \citep{park2026vrh}, retrieve evidence in long multimodal contexts \citep{li2026mmretrieval}, point to and write out the image relevant to a question among several \citep{park2026write}, and, in \qwen{8B}, track the region being described, so that redirecting their attention steers the description \citep{gandikota2026gaze}. Multiple-choice answering involves symbol-producing heads \citep{wiegreffe2025mcqa} and a decide-then-bind sequence \citep{wong2026decide}. These studies concern text-to-image, within-image, or whole-image retrieval; we study region-to-region comparison \emph{between} two images. Interventions have known artifacts: conclusions depend on the counterfactual and metric \citep{zhang2024patching}, subspace patching can activate dormant pathways \citep{makelov2024subspace}, patched states can leave the natural distribution \citep{grant2026divergent}, and mid-layer interventions in VLMs can yield null results \citep{zhang2026disconnected}. We pair each intervention with random-component controls and a residual-norm check.

\paragraph{Mechanism-guided repair.}
Training-free repairs gate causal routes \citep{cheng2026crg}, reweight attention \citep{liu2026seeing}, edit sensitivity-selected heads \citep{ji2026causallens}, or attenuate a language-prior subspace \citep{zhuang2026mgap}. Parameter-efficient fine-tuning \citep{hu2022lora} has been restricted to visual tokens and selected attention paths \citep{luo2025imagelora} and to heads selected by their mechanistic role \citep{mitra2025robotic}. We restrict adaptation to heads selected by a causal comparison criterion, compare the result against head-matched and parameter-matched alternatives, and test whether unconstrained adaptation converges on the same heads.

\section{Setup}
\label{sec:setup}

\paragraph{Tasks.}
Every item is a four-way multiple-choice question over two images: image~1 contains a location labelled \refm{}; image~2 contains four candidates labelled A--D, one of which corresponds to \refm{}. In \syn{}-squiggle, shapes are closed cubic splines through $n{=}30$ anchors; \syn{}-known uses ten nameable outline classes matched in size and stroke. In both, image~2 contains the same four shapes as image~1 at new positions, identical up to translation, so every option has a copy in image~1 and only \refm{}-specific comparison solves the task; within-item shape pairs are separated by a minimum Chamfer distance. \hp{} is same-instance point correspondence generated from HPatches homographies \citep{balntas2017hpatches}; \spt{} is semantic keypoint correspondence from SPair-71k \citep{min2019spair}. Distractors are at least 8\% of the image diagonal from the target and from each other. Splits are by seed range (\syn{}), sequence (\hp{}), or official split (\spt{}), with 1{,}000 development and 1{,}000 test items per task (500 each for \syn{}-known) unless stated. For transfer we also evaluate the BLINK semantic- and visual-correspondence validation splits \citep{fu2024blink}, each with fewer than 200 items.

\paragraph{Models.}
\qwen{2B/4B/8B/32B} \citep{bai2025qwen3vl}, \ivl{} \citep{wang2025internvl35}, and \gemma{} \citep{gemmateam2025gemma3}. Qwen3-VL uses SigLIP2-Large for 2B/4B and SigLIP2-SO-400M for 8B/32B, so scale comparisons are made only within encoder-matched pairs. For \qwen{8B} we verified that perturbing image~1 leaves image~2's encoder output unchanged, so cross-image comparison can occur only in the language model. Mechanistic analyses and repairs use \qwen{8B} and \ivl{}.

\paragraph{Scoring.}
One forward pass returns the log-probabilities of the four letter tokens at the scoring position, with the final normalization and unembedding computed in fp32. The answer format was chosen on development data so that nearly all probability mass falls on the four letters and greedy generation agrees with the highest-scoring letter (Appendix~\ref{app:repro}); it differs from the prefill format of \citet{shahgir2026words}, so absolute accuracies are not directly comparable with theirs. Each item is scored under four cyclic letter orders that keep shape positions fixed.

\paragraph{Readouts.}
We read the signal at four points. $\Acc$ is behavioural accuracy, and $\Raff$ is accuracy after a multinomial logistic calibration of the four letter log-probabilities, fitted on development data; together they bound what the output already carries. $\Pans$ is an input-matched probe on the answer-position residual stream (PCA to 256 dimensions, multinomial logistic regression; layer and $\ell_2$ strength by five-fold cross-validation on development data): it sees exactly the positions the model uses to answer. $\Preg$ is a privileged readout over annotated regions, the maximum of a training-free $\MaxSim$ score and a paired probe $\Ppair$ (the latter counted only with selectivity $\ge 5\pp$ over shuffled labels), with layers chosen on development data; it measures what the visual tokens hold. Confidence intervals are 2{,}000-sample item bootstraps. Appendix~\ref{app:notation} collects all symbols.

\paragraph{Evaluation sets.}
Three disjoint groups of test items are used: the measurement test sets of \S\ref{sec:results} (letter order~0), the held-out confirmation items of \S\ref{sec:res_sel}, and 1{,}000 further \spt{} test pairs for the repair experiments of \S\ref{sec:repair} (mean over the four letter orders, or order~0 where stated). Every comparison we report is paired within a single item set, so base accuracies vary slightly across sections (\qwen{8B} on \spt{}: $.438$, $.430$, and $.436$).

\paragraph{Reproducibility.}
Thresholds and decision criteria were fixed before each analysis, and those that were not met are listed in Appendix~\ref{app:criteria}; every reported number was recomputed from raw per-item outputs by independent analysis code (Appendix~\ref{app:repro}). Region-to-token mappings were validated per family by a patch-level perturbation test (precision $1.0$). The full study used about 90 GPU-hours, mainly on a single RTX PRO 6000. Code, item generators, split identifiers, head indices, and trained adapters will be released.

\section{Where the Signal Is Lost}
\label{sec:results}

\subsection{Privileged Readouts and Nameability}
\label{sec:res_ladder}

We begin with exact copies, where the right answer is fixed by pixel identity, and ask how much of it each readout recovers.

\paragraph{Privileged readouts saturate on exact copies.}
On \syn{}-squiggle (test), \qwen{8B} answers 41\% correctly while $\Pans$ reaches 64\% and $\MaxSim$/$\Ppair$ reach $1.00$; \qwen{2B} answers 31\% with $\Pans=36\%$ and $\MaxSim/\Ppair=0.93/1.00$. The privileged readouts are near ceiling already at the vision-encoder output, because the reference and the target are pixel-identical. On this task they measure the input, and the informative quantity is the input-matched gap $\Pans-\Acc$.

\paragraph{Nameability does not explain the gap.}
The gap $\Pans-\Acc$ is nearly the same for squiggles and nameable shapes, differing by $+1.8\pp$ for \qwen{8B} and $+2.8\pp$ for \qwen{2B}. Naming raises both $\Pans$ and $\Acc$ and leaves the distance between them unchanged.

\paragraph{The \qwen{8B} gap survives calibration.}
Calibrating the letter log-probabilities leaves the \qwen{8B} gap intact on both conditions, while it closes the small \qwen{2B} gap on known shapes (Table~\ref{tab:calib}). The two halves of the task are each solved: for known shapes, \qwen{8B} names the \refm{} shape and finds the letter of a named shape (letter binding 100\%). Yet it answers the combined question correctly on 63\% of items, below the product of the two step accuracies ($\approx 82\%$).

\begin{table}[t]
\centering
\caption{\textbf{Calibration does not close the \qwen{8B} gap.} $\Pans-\Raff$ on \syn{} test, order 0, with paired 95\% CIs.}
\label{tab:calib}
\begin{tabular}{llc}
\toprule
Model & Condition & $\Pans-\Raff$ (pp) \\
\midrule
\qwen{2B} & squiggle & $+6.0$ [$+2.2$, $+9.5$] \\
\qwen{2B} & known & $-0.2$ [$-5.0$, $+4.4$] \\
\qwen{8B} & squiggle & $+21.3$ [$+17.9$, $+24.6$] \\
\qwen{8B} & known & $+19.0$ [$+14.8$, $+23.0$] \\
\bottomrule
\end{tabular}
\end{table}

\subsection{Breadth: The Readout Gap Is Specific to Exact Copies}
\label{sec:res_breadth}

Across families, the calibration-robust gap on \syn{}-squiggle persists ($+15$ to $+25\pp$ in all models except \qwen{2B}). Real correspondence behaves differently (Table~\ref{tab:real}): calibration changes $\Acc$ by at most $2.6\pp$ across all six models, and $\Pans$ is within a few points of $\Raff$ in every model. \hp{} is solved well above chance (answer-position probe $0.54$--$0.86$), whereas \spt{} is near chance both at the answer position and in the output. With both images replaced by grey, the transferred probe stays at or below $0.315$. Within encoder-matched pairs, the \syn{} gap grows from 2B to 4B and shrinks from 8B to 32B (where known shapes reach ceiling); it is stable across scale on \hp{} and \spt{}.

\paragraph{The gap is not a property of the probe.}
Every task and model uses the same probe architecture, dimensionality, and model-selection protocol, with layer and regularization chosen on that task's development data. If the probe's extra capacity relative to a calibrated four-way output were enough to create a gap, the gap would appear everywhere. It does not. On \hp{} and \spt{} the probe exceeds $\Raff$ by at most $2.6\pp$ and falls below it in three of six model--task cells, and on \syn{}-known calibration closes the \qwen{2B} gap entirely, yet for \qwen{8B} the same protocol exceeds $\Raff$ by about $20\pp$ on both \syn{} conditions. The \syn{} gap is specific to exact copies in the models that compare them strongly (\S\ref{sec:res_sel}).

\paragraph{The decoded answer is not on the output path.}
To test whether the output uses the answer-position signal, we edit the residual stream at the answer position and the probe's layer, along the probe's own three-dimensional subspace $S$. We either remove the component in $S$ or scale it by $\gamma\in\{2,4\}$, and compare with 20 random subspaces matched in edit magnitude; no label information enters the edit. Neither edit moves the answer: on \syn{}-squiggle and on \hp{}, in both \qwen{8B} and \ivl{}, $\Acc$ changes by at most $0.2\pp$, within the range of the random subspaces, and residual norms stay in the natural band (Appendix~\ref{app:edits}). The layer itself is responsive. Adding the output direction of a letter the model did not choose, at the same position and layer and at eight times the natural scale of $S$, switches the answer to that letter on $2.8$--$3.8\%$ of \syn{} and $9.6$--$18.6\%$ of \hp{} items, compared with at most $0.4\%$ and $1.8\%$ for ten random directions of equal norm. Steering along the probe's class direction, toward a class it did not predict, moves at most $1.6\%$ of the items, within the range of the random directions (Appendix~\ref{app:controls}). The answer that the probe reads is therefore not the one the model reports. On \hp{}, where the output is mostly correct, the model reaches its answer through other channels. On \syn{}, the decodable comparison signal lies off the output path, in line with reports that decodable signals in VLMs need not be causal for the answer \citep{zhang2026disconnected,veikherman2026encoded}. Because amplifying it at a single layer leaves the output unchanged, the loss lies beyond a weak linear readout.

\begin{table}[t]
\centering
\caption{\textbf{Real correspondence shows no hidden answer-position gap.} Test, order 0, $n=1{,}000$.}
\label{tab:real}
\small\setlength{\tabcolsep}{2.6pt}
\begin{tabular}{llcccc}
\toprule
Model & Task & $\Acc$ & $\Raff$ & $\Pans$ & \begin{tabular}[b]{@{}c@{}}$\Pans-\Raff$\\{}[95\% CI]\end{tabular} \\
\midrule
\multirow{2}{*}{Qwen-8B} & \hp{} & .770 & .778 & .804 & $+2.6$ [$+0.6$, $+4.8$] \\
 & \spt{} & .438 & .460 & .433 & $-2.7$ [$-6.1$, $+0.7$] \\
\multirow{2}{*}{InternVL-8B} & \hp{} & .543 & .536 & .542 & $+0.6$ [$-2.3$, $+3.2$] \\
 & \spt{} & .355 & .369 & .310 & $-5.9$ [$-9.7$, $-2.1$] \\
\multirow{2}{*}{Gemma-12B} & \hp{} & .590 & .591 & .601 & $+1.0$ [$-1.7$, $+3.7$] \\
 & \spt{} & .407 & .404 & .351 & $-5.3$ [$-8.8$, $-1.8$] \\
\bottomrule
\end{tabular}
\end{table}

\subsection{Semantic Correspondence Is Present in the Visual Tokens}
\label{sec:res_priv}

If semantic correspondence never reaches the answer position, the question is whether it exists anywhere in the language model. We compute the region readout $\Preg$ over $30\times30$-pixel regions centred on each point, on images without drawn markers, and compare it with the answer-position probe (Table~\ref{tab:priv}). On \spt{}, the region readout exceeds the answer-position probe in all three models: by $+16.3\pp$ for \qwen{8B}, $+25.9\pp$ for \ivl{}, and $+14.3\pp$ for \gemma{}. In no cell is $\Preg$ attained only at the vision-encoder output, so the language model's own visual tokens carry the correspondence, in line with \citet{liu2025visualrep}. Selecting the readout layer on the test set instead of development data changes $\MaxSim$ by $0$--$1.8\pp$, with every interval including zero.

\begin{table}[t]
\centering
\caption{\textbf{Region readout minus answer-position probe}, $\Preg-\Pans$ in pp (images without markers, test).}
\label{tab:priv}
\begin{tabular}{lccc}
\toprule
Model & \syn{}-squiggle & \hp{} & \spt{} \\
\midrule
\qwen{8B} & $+36.0$ & $+5.8$ & $+16.3$ \\
\ivl{} & $+25.8$ & $+37.0$ & $+25.9$ \\
\gemma{} & $+35.1$ & $+18.1$ & $+14.3$ \\
\bottomrule
\end{tabular}
\end{table}

\paragraph{Marker control.}
When everything except the drawn markers is replaced by grey, the region readout falls $5.8$--$16.1\pp$ below its value on marked images in five of six model--task cells; in \ivl{} on \hp{} the markers alone carry most of it ($0.780$ versus $0.804$). Table~\ref{tab:priv} therefore uses images without markers, so that the readout reflects image content.

\section{How the Two Images Are Compared}
\label{sec:heads}

\subsection{Cross-Image Heads}
\label{sec:res_heads}

Image-1 tokens precede image-2 tokens, so under causal attention a cross-image comparison must run from image-2 (option) queries to image-1 keys, or through later positions. This constraint makes the comparison searchable head by head. We scored each head by how strongly option tokens attend to their corresponding region in image~1 on \syn{} development items and blocked the image-2$\to$image-1 attention of the top-$k$ heads, with $k$ chosen on development data. This gives a set $C_{32}$ of 32 heads in layers 1--24 of \qwen{8B} and a set $C_{16}$ of 16 heads in layers 3--23 of \ivl{}; we write $C_k$ for either.

\paragraph{Knockouts that preserve attention mass.}
Blocking this attention lowers $\Acc$ on the 1{,}000 \syn{}-squiggle test items by $15.2\pp$ for \qwen{8B} and $24.2\pp$ for \ivl{}, more than the 95th percentile of 20 layer-matched random head sets, while blocking the same heads' attention within image~2 has a much smaller effect. Blocking also removes attention mass: for \qwen{8B}, 53\% of these items leave the base model's residual-norm range (1st--99th percentile; \ivl{}: 1.9\%), and on the 500 items of Table~\ref{tab:knock}, where 49\% leave, the drop among the items that stay in range is only $3.1\pp$. To isolate \emph{where} the attention goes from \emph{how much} there is, we use two operations that keep the total attention of each image-2 query on image~1: \textsc{Uniform} spreads it evenly over the image-1 tokens, and \textsc{Shuffle} permutes it among them. We apply each operation to $C_k$, to ten layer-matched random head sets, and to the three attention-matched head sets of \S\ref{sec:res_repair} (Table~\ref{tab:knock}).

On \syn{} and \hp{}, both operations lower $\Acc$ by $15$--$32\pp$ and $\Pans$ by $20$--$48\pp$ in both models. For every operation and both measures, $C_k$ ranks first among the ten random sets, and its $\Acc$ drop exceeds that of all three attention-matched sets, none of which lowers $\Acc$ by more than $6.8\pp$. Both operations keep $94$--$98\%$ of the \syn{} items in the norm range, and on these items the drops are unchanged ($18.0$ and $19.3\pp$ for \qwen{8B}). On \hp{}, $6$--$16\%$ of items leave the range under these two operations ($21\%$ under \textsc{Block}), and the drops on in-range items are within $2\pp$ of the overall drops. On \spt{}, the knockouts change accuracy by at most $3.6\pp$ in \qwen{8B} and not measurably in \ivl{}: the base models' semantic answers do not yet run through these heads. This is precisely the gap that the repair of \S\ref{sec:repair} targets.

\begin{table}[t]
\centering
\caption{\textbf{Knockouts of the image-2$\to$image-1 attention of $C_k$.} Drop in $\Acc$ (pp) on 500 test items per task; in parentheses, the drop on items whose residual norm stays in the base band. \textsc{Uniform} and \textsc{Shuffle} preserve attention mass; \textsc{Block} removes it (on these 500 items; the $15.2$ and $24.2\pp$ above are on all 1{,}000 \syn{} test items). Ctrl: largest drop of any of ten random and three attention-matched head sets under any operation.}
\label{tab:knock}
\small\setlength{\tabcolsep}{2.2pt}
\begin{tabular}{llccccc}
\toprule
 & Task & Base & \textsc{Uniform} & \textsc{Shuffle} & \textsc{Block} & Ctrl \\
\midrule
\multicolumn{7}{l}{\emph{\qwen{8B}}} \\
 & \syn{} & 41.2 & 17.2 (18.0) & 18.2 (19.3) & 17.4 (\phantom{0}3.1) & 3.6 \\
 & \hp{} & 76.2 & 32.2 (31.8) & 31.8 (31.4) & 21.2 (20.9) & 2.4 \\
 & \spt{} & 46.6 & \phantom{0}2.2 (\phantom{0}2.4) & \phantom{0}3.2 (\phantom{0}3.7) & \phantom{0}3.6 (\phantom{0}3.8) & 2.0 \\
\midrule
\multicolumn{7}{l}{\emph{\ivl{}}} \\
 & \syn{} & 48.6 & 20.2 (19.9) & 22.0 (22.0) & 22.4 (22.2) & 6.8 \\
 & \hp{} & 53.6 & 16.8 (16.2) & 14.6 (12.6) & 16.2 (14.9) & 4.2 \\
 & \spt{} & 34.8 & $-$0.8 ($-$0.8) & $-$0.6 ($-$0.6) & $-$0.6 ($-$0.6) & 1.0 \\
\bottomrule
\end{tabular}
\end{table}

\paragraph{Where the heads look.}
On \hp{} and \spt{}, the true option's attention peaks on the \refm{} region in 94--100\% of items. Distractors also attend mostly to \refm{} rather than to their own correspondence, so per-option retrieval accuracy is $0.28$--$0.33$. Since \refm{} is the only marked location in image~1, this pattern points to two ingredients, salience and content, which we separate next. Figure~\ref{fig:qual} shows both on representative items (selection in Appendix~\ref{app:repro}).

\begin{figure*}[t]
\centering
\includegraphics[width=\textwidth]{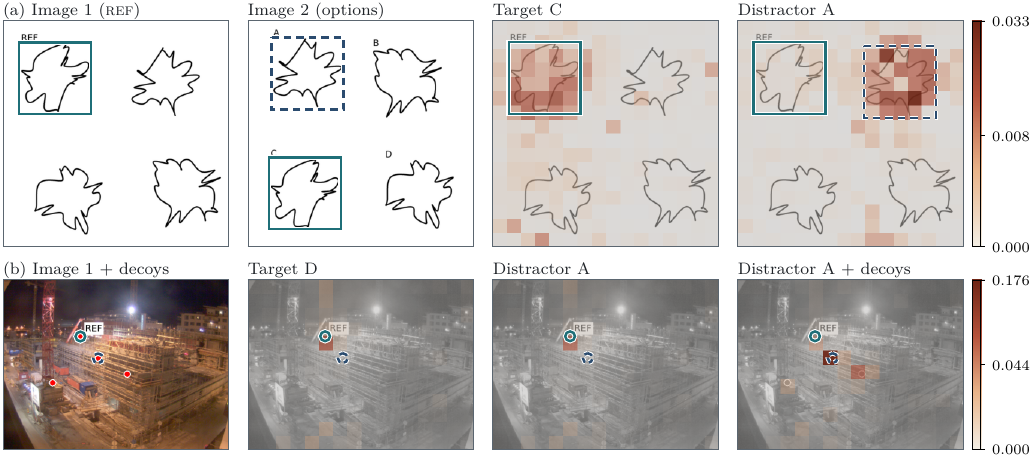}
\caption{\textbf{What the cross-image heads attend to} (\qwen{8B}, mean over the 32 heads of $C_k$, held-out items). Each heatmap shows the attention from one option's tokens in image~2 to each token of image~1, on a square-root colour scale shared within a row. Teal: \refm{}; dashed blue: the distractor's own corresponding location. \textbf{(a)} Exact copies: target C places 53\% of its image-1 attention on \refm{}; distractor A places 6\% there and 55\% on its own copy. \textbf{(b)} \hp{}: without decoys, distractor A attends to the salient \refm{} marker as the target D does (38\% of its image-1 attention, 0.1\% on its own location). With identical unlabelled markers drawn at the distractors' locations, its attention moves to its own marked location (36\%, against 8\% on \refm{}), and the model's answer switches to A. Repair items are shown in Appendix~\ref{app:maps}.}
\label{fig:qual}
\end{figure*}
\subsection{Content-Selective, Salience-Gated Attention}
\label{sec:res_sel}

On held-out items not used in any earlier analysis (600 \syn{}, 474 \hp{}, 400 \spt{}), with the head sets fixed from \S\ref{sec:res_heads}, we measure, averaged over the heads in $C_k$, the attention mass $m_{\refm}(o)$ from option $o$ to the \refm{} region, the selectivity $\dref = m_{\refm}(\text{target}) - \overline{m_{\refm}(\text{distractor})}$, and the ratio of means $\rref$.

\begin{table}[t]
\centering
\caption{\textbf{Selectivity of cross-image heads on held-out test items.} AUC uses $\dref$ as a score for answer-position probe correctness.}
\label{tab:sel}
\setlength{\tabcolsep}{3pt}
\begin{tabular}{llccc}
\toprule
Model & Task & $\dref$ [95\% CI] & $\rref$ & AUC \\
\midrule
\multirow{3}{*}{Qwen-8B} & \syn{} & $+.252$ [$.244$, $.260$] & 10.36 & .48 \\
 & \hp{} & $+.051$ [$.046$, $.056$] & 1.88 & .70 \\
 & \spt{} & $+.011$ [$.008$, $.013$] & 1.30 & .66 \\
\multirow{3}{*}{InternVL-8B} & \syn{} & $+.155$ [$.148$, $.163$] & 2.88 & .69 \\
 & \hp{} & $+.046$ [$.035$, $.057$] & 1.32 & .62 \\
 & \spt{} & $+.009$ [$.003$, $.016$] & 1.13 & .56 \\
\bottomrule
\end{tabular}
\end{table}

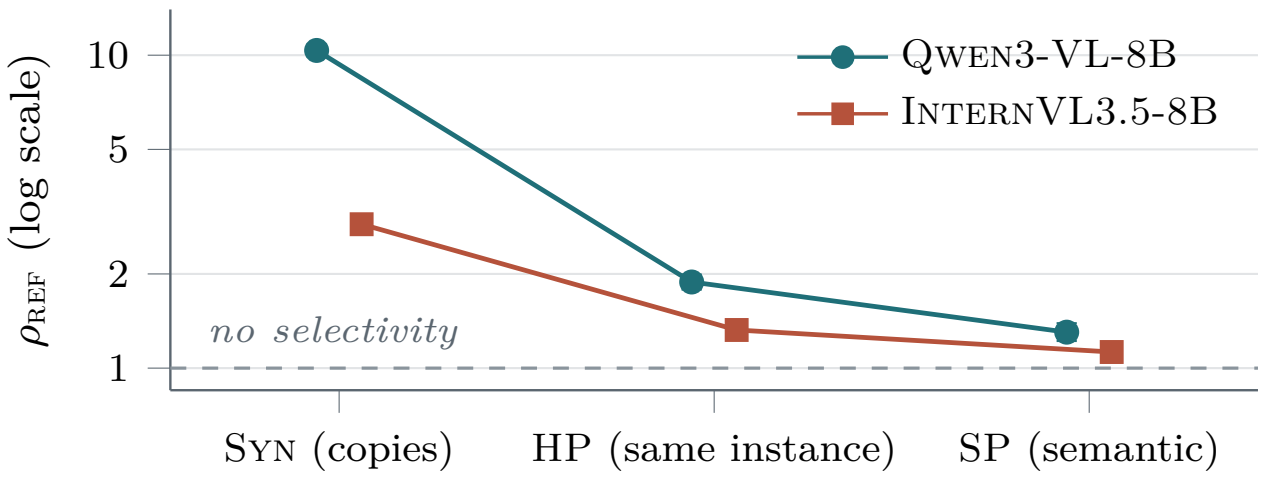
\begin{figure}[t]
\centering
\begin{tikzpicture}
\begin{axis}[
  width=\columnwidth, height=4.1cm, scale only axis=false,
  ymode=log, ymin=.85, ymax=14,
  ytick={1,2,5,10}, yticklabels={1,2,5,10},
  xmin=.55, xmax=3.45, xtick={1,2,3},
  xticklabels={\syn{} (copies),\hp{} (same instance),\spt{} (semantic)},
  ylabel={$\rref$ (log scale)},
  axis x line*=bottom, axis y line*=left,
  axis line style={draw=sbuSlate,line width=.45pt},
  tick style={draw=sbuSlate}, tick align=outside,
  ymajorgrids=true, grid style={draw=sbuLine!25},
  legend style={draw=none,fill=none,at={(.99,.98)},anchor=north east},
  every axis plot/.append style={line width=.9pt},
]
\addplot[sbuLine,dashed,line width=.6pt,forget plot] coordinates {(.55,1)(3.45,1)};
\node[anchor=south west,font=\rmfamily\itshape\fontsize{7}{8}\selectfont,text=sbuSlate] at (axis cs:.6,1) {no selectivity};
\addplot[sbuTeal,mark=*,mark size=1.9pt,
  error bars/.cd,y dir=both,y explicit,error bar style={line width=.6pt}]
  coordinates {(0.94,10.356) += (0,0.409) -= (0,0.381)
               (1.94,1.885) += (0,0.107) -= (0,0.103)
               (2.94,1.303) += (0,0.089) -= (0,0.082)};
\addplot[sbuBrick,mark=square*,mark size=1.8pt,
  error bars/.cd,y dir=both,y explicit,error bar style={line width=.6pt}]
  coordinates {(1.06,2.883) += (0,0.127) -= (0,0.123)
               (2.06,1.324) += (0,0.087) -= (0,0.082)
               (3.06,1.126) += (0,0.087) -= (0,0.084)};
\legend{\qwen{8B},\ivl{}}
\end{axis}
\end{tikzpicture}
\caption{\textbf{Selectivity falls from copies to semantic correspondence.} Ratio of the true option's attention on the reference to that of distractors, averaged over $C_k$, on held-out test items with 95\% intervals. The \spt{}$-$\hp{} difference in $\dref$ is $-0.040$ [$-0.046$, $-0.034$] for \qwen{8B} and $-0.036$ [$-0.049$, $-0.024$] for \ivl{}.}
\label{fig:sel}
\end{figure}

\paragraph{Selectivity, its drop, and prediction.}
In both models the true option attends more to the reference than distractors do ($\dref>0$ on \hp{} and \spt{}). Selectivity falls from \hp{} to \spt{}, in both the difference and the ratio (Fig.~\ref{fig:sel}), mirroring the drop in accuracy, and $\dref$ predicts whether the answer-position probe is correct, with AUC $0.62$--$0.70$ on \hp{} (Table~\ref{tab:sel}). Removing items whose correspondence regions overlap leaves these values essentially unchanged.

\paragraph{Salience gating.}
When identical unlabelled markers are drawn at the distractors' correspondence locations in image~1, distractors shift their attention to their own, now marked, locations (ratio of attention on \refm{} to attention on the own location: $0.19$--$0.77$), and selectivity remains positive and becomes larger ($\dref=+0.109$ on \hp{} for \qwen{8B}). On single items the shift is visible directly (Fig.~\ref{fig:qual}b): a distractor's attention moves from \refm{} to its own location once that location is marked, and the answer can follow it. Without such markers, distractors place almost no attention on their own correspondence ($0.011$--$0.017$). Attention therefore goes first to salient candidate locations in image~1 and is then weighted by content similarity among them. This division of labour fits causal attention: image~1 is encoded before any candidate is seen, so a visible marker is the cue an option can use to narrow its search, and content then decides among the locations that pass this filter.

\paragraph{The marker sets the target; content still pulls.}
Moving the \refm{} marker to a distractor's correspondence location, with image~2 unchanged, makes the new target selective ($\dref=+0.052$ on \hp{} for \qwen{8B}), while the old target keeps a smaller residual ($+0.009$) from content-driven attention to its now unmarked correspondence. Selectivity thus follows the marker, and content still pulls attention toward the true match.

\paragraph{Similarity, not temperature.}
Multiplying the image-2$\to$image-1 attention logits of $C_k$ by $\beta\in\{1.5,2\}$ does not raise calibrated accuracy on \hp{} or \spt{} in either model, and on \syn{} it lowers the answer-position probe. If the comparison stage limits semantic correspondence, what needs to change is the similarity these heads compute rather than its temperature. We test this next.

\section{Repairing the Comparison}
\label{sec:repair}

If the cross-image heads limit semantic correspondence, changing \emph{what they compare} should help, and should help more than changing other components of similar size. We test this with adapters restricted to $C_k$ and with matched alternatives, and then ask whether adapters that are free to change any layer converge on the same heads.

\subsection{Targeted Adapters}
\label{sec:res_repair}

\paragraph{Arms.}
All arms share one recipe and differ only in their trainable parameters, with base weights frozen. \textbf{A} trains rank-16 LoRA updates \citep{hu2022lora} on the query and key projections of the heads in $C_k$ (for grouped-query attention, the key update acts on the key--value group of each selected head): 1.68M parameters for \qwen{8B} and 1.50M for \ivl{}. \textbf{A$_Q$} trains the query updates only (0.85M / 0.75M); because query projections are separate per head, A$_Q$ changes the computation of the selected heads and of no other head. \textbf{B} applies arm A to three layer-matched random head sets of the same size. \textbf{C} trains the value projections of $C_k$ with nearly the same parameter count as A (1.65M). \textbf{D} is a standard LoRA on all attention projections of all layers at rank 2 (1.92M parameters, the closest match to A). \textbf{E} is a rank-16 LoRA on all attention projections of all layers (15.3M), a common default that is not parameter-matched. C and E were run for \qwen{8B} only.

\paragraph{Training and evaluation.}
Training uses 2{,}000 items from the SPair-71k training split in the same four-option format, cross-entropy on the correct letter at the scoring position, AdamW for two epochs with batch size 8 and a cosine schedule; the learning rate is chosen per arm from $\{10^{-4},3\cdot10^{-4},10^{-3}\}$ on 400 validation items. A, C, and D use three seeds. The primary test set is 1{,}000 \spt{} test pairs never used in earlier runs, scored as the mean over the four letter orders. Transfer is measured on the held-out \hp{} items of \S\ref{sec:res_sel} and on BLINK semantic and visual correspondence, none of which is used for training or selection; the two \syn{} conditions serve as collateral checks.

\begin{table*}[t]
\centering
\caption{\textbf{Repair arms.} Accuracy (mean over four letter orders; seeds averaged) after training on 2{,}000 \spt{} items. Only \spt{} is in-domain; \hp{} and BLINK measure transfer, \syn{} measures collateral effects. Best value per column and model in bold.}
\label{tab:repair}
\setlength{\tabcolsep}{4.2pt}
\begin{tabular}{lrcccccc}
\toprule
Arm (trainable parameters) & Params & \spt{} & \hp{} & BLINK-SC & BLINK-VC & \syn{}-squiggle & \syn{}-known \\
\midrule
\multicolumn{8}{l}{\emph{\qwen{8B}}} \\
base (none) & -- & .430 & .782 & .367 & .669 & .414 & .686 \\
A: QK of $C_{32}$ & 1.68M & .712 & .934 & .549 & .802 & \textbf{.551} & .833 \\
A$_Q$: Q of $C_{32}$ & 0.85M & .674 & .922 & .504 & .791 & .480 & .842 \\
B: QK of random heads & 1.69M & .677 & .899 & .499 & .808 & .451 & .701 \\
C: V of $C_{32}$ & 1.65M & .747 & .925 & .561 & .876 & .488 & .785 \\
D: all layers, rank 2 & 1.92M & \textbf{.792} & .953 & \textbf{.655} & .899 & .402 & .806 \\
E: all layers, rank 16 & 15.3M & .790 & \textbf{.958} & \textbf{.655} & \textbf{.907} & .451 & \textbf{.911} \\
\midrule
\multicolumn{8}{l}{\emph{\ivl{}}} \\
base (none) & -- & .354 & .531 & .309 & .436 & \textbf{.526} & .926 \\
A: QK of $C_{16}$ & 1.50M & .605 & .837 & .367 & .562 & .366 & .776 \\
A$_Q$: Q of $C_{16}$ & 0.75M & .581 & .836 & .410 & .558 & .315 & .784 \\
B: QK of random heads & 1.51M & .584 & .829 & .369 & .622 & .474 & .906 \\
D: all layers, rank 2 & 1.92M & \textbf{.728} & \textbf{.941} & \textbf{.614} & \textbf{.845} & .492 & \textbf{.993} \\
\bottomrule
\end{tabular}
\end{table*}

\paragraph{Training the comparison heads repairs most of the deficit.}
Arm A raises \spt{} accuracy by $+28.3\pp$ [$25.3$, $31.1$] for \qwen{8B} (.430$\to$.712) and by $+25.1\pp$ [$22.4$, $27.8$] for \ivl{} (.354$\to$.605), while training about $0.02\%$ of the model's parameters (Table~\ref{tab:repair}). The gains remain large with only 500 training items ($+23.9$ and $+19.0\pp$). Training the query projections alone (A$_Q$), which affects no head outside $C_k$, gives $86$--$90\%$ of the effect ($+24.4$ and $+22.7\pp$), so the repair resides in the selected heads themselves rather than in keys shared with other heads. The gains carry over to HPatches and to both BLINK correspondence splits, none of which was used for training (Table~\ref{tab:repair}); the one transfer gain whose interval includes zero is BLINK semantic correspondence for \ivl{} ($+5.8\pp$). For \qwen{8B}, A also improves both \syn{} conditions.

\paragraph{Head identity matters beyond layer.}
Adapting the comparison heads beats adapting the same number of random heads in the same layers, by $+3.6\pp$ for \qwen{8B} and $+2.1\pp$ for \ivl{}, with 95\% intervals above zero, also when the random head sets are resampled together with the items (Table~\ref{tab:repaircmp}). The advantage holds, and grows for \ivl{}, when only query projections are trained, so that no update can reach a head outside the chosen set. In that case A$_Q$ beats query-only training of the same random sets (B$_Q$) by $+3.5\pp$ and $+4.5\pp$, again with intervals above zero. Query updates of random heads leave the selectivity of $C_k$ almost unchanged ($\dref=.013$--$.016$ and $.010$--$.011$, against $.034$ and $.029$ for A$_Q$).

\paragraph{Attention mass does not explain the advantage.}
Random heads in the same layers send only $32$--$47\%$ as much attention across images as $C_k$, so A$-$B could reflect access to the other image rather than the role of the heads. We therefore built three further head sets per model, drawn from any layer and matched to $C_k$ head by head in cross-image attention mass on \syn{} development items ($80$--$99\%$ of the mass of $C_k$; $1.3$--$1.7$ times as much on \spt{}), while excluding the heads with the highest correspondence scores. We trained them with the recipes of A (arm M) and A$_Q$ (arm M$_Q$), with every matched arm having more trainable parameters than its reference. Matched heads do about as well as random ones (\spt{} $.679$ vs.\ $.677$ for \qwen{8B} and $.580$ vs.\ $.584$ for \ivl{}), and $C_k$ beats them, by $+3.3\pp$ and $+2.5\pp$ (A$-$M) and $+2.1\pp$ and $+4.3\pp$ (A$_Q-$M$_Q$), all four with intervals above zero (Table~\ref{tab:matched}). The advantage of $C_k$ is therefore explained neither by layer (B) nor by attention mass (M); it lies in the content of their attention. The matched arms also show that these heads serve both kinds of comparison. In \ivl{}, training M leaves \syn{}-squiggle and \syn{}-known at $.473$ and $.912$, whereas retraining $C_k$ for semantic correspondence lowers them to $.366$ and $.776$ (Table~\ref{tab:repair}), in line with the causal role of the same heads in exact-copy comparison (\S\ref{sec:res_heads}).

\paragraph{Unconstrained adapters reach higher accuracy.}
The parameter-matched all-layer adapter D exceeds A by $8.0\pp$ (\qwen{8B}) and $12.3\pp$ (\ivl{}), and adapting the value projections of the same heads (C) exceeds adapting their query--key projections by $3.5\pp$ in \qwen{8B}. These adapters can change the features being compared as well as the comparison itself. Section~\ref{sec:convergence} shows that the attention of $C_k$ alone nonetheless carries a large share of the all-layer adapters' gain when transplanted into the base model.

\begin{figure*}[t]
\centering
\begin{tikzpicture}
\begin{groupplot}[
  group style={group size=2 by 1, horizontal sep=1.55cm},
  height=5.0cm,
  axis x line*=bottom, axis y line*=left,
  axis line style={draw=sbuSlate,line width=.45pt},
  tick style={draw=sbuSlate}, tick align=outside,
  grid style={draw=sbuLine!25},
]
\nextgroupplot[
  width=.43\textwidth,
  xmin=0, xmax=.088, ymin=.32, ymax=.84,
  xtick={0,.02,.04,.06,.08},
  xticklabels={0,.02,.04,.06,.08},
  ytick={.4,.5,.6,.7,.8},
  xlabel={Selectivity $\dref$ of $C_k$ on \spt{} test},
  ylabel={\spt{} test accuracy},
  ymajorgrids=true, xmajorgrids=true,
  legend style={draw=none,fill=none,at={(.98,.03)},anchor=south east},
  title={\textbf{(a)} Selectivity tracks the repair across arms},
  title style={font=\rmfamily\fontsize{8.4}{10}\selectfont,yshift=-2pt},
]
\addplot[only marks,mark=*,mark size=2.1pt,draw=sbuTeal,fill=sbuTeal]
  coordinates {(.0107,.430) (.0426,.712) (.0248,.677) (.0477,.747) (.0771,.792)};
\addlegendentry{\qwen{8B}}
\addplot[only marks,mark=square*,mark size=1.9pt,draw=sbuBrick,fill=sbuBrick]
  coordinates {(.0084,.354) (.0298,.605) (.0156,.584) (.0698,.728)};
\addlegendentry{\ivl{}}
\addplot[only marks,mark=*,mark size=2.1pt,draw=sbuTeal,fill=white,line width=.8pt,forget plot]
  coordinates {(.0342,.674) (.0711,.790)};
\addplot[only marks,mark=square*,mark size=1.9pt,draw=sbuBrick,fill=white,line width=.8pt,forget plot]
  coordinates {(.0293,.581)};
\addplot[only marks,mark=diamond*,mark size=2.6pt,draw=sbuTeal,fill=sbuTeal!35,line width=.6pt,forget plot]
  coordinates {(.0229,.679) (.0164,.652)};
\addplot[only marks,mark=diamond*,mark size=2.6pt,draw=sbuBrick,fill=sbuBrick!35,line width=.6pt,forget plot]
  coordinates {(.0146,.580) (.0121,.537)};
\begin{scope}[font=\rmfamily\fontsize{7}{8}\selectfont,inner sep=1.4pt]
\node[text=sbuTeal!80!black,anchor=south] at (axis cs:.0107,.437) {base};
\node[text=sbuTeal!80!black,anchor=south east] at (axis cs:.0420,.716) {A};
\node[text=sbuTeal!80!black,anchor=north west] at (axis cs:.0348,.669) {A$_Q$};
\node[text=sbuTeal!80!black,anchor=south west] at (axis cs:.0250,.682) {B};
\node[text=sbuTeal!80!black,anchor=south] at (axis cs:.0477,.752) {C};
\node[text=sbuTeal!80!black,anchor=south west] at (axis cs:.0775,.796) {D};
\node[text=sbuTeal!80!black,anchor=south east] at (axis cs:.0706,.795) {E};
\node[text=sbuBrick!85!black,anchor=west] at (axis cs:.0098,.354) {base};
\node[text=sbuBrick!85!black,anchor=south west] at (axis cs:.0304,.609) {A};
\node[text=sbuBrick!85!black,anchor=north west] at (axis cs:.0300,.577) {A$_Q$};
\node[text=sbuBrick!85!black,anchor=south] at (axis cs:.0160,.590) {B};
\node[text=sbuBrick!85!black,anchor=north west] at (axis cs:.0704,.724) {D};
\node[text=sbuTeal!80!black,anchor=north] at (axis cs:.0229,.672) {M};
\node[text=sbuTeal!80!black,anchor=east] at (axis cs:.0153,.652) {M$_Q$};
\node[text=sbuBrick!85!black,anchor=east] at (axis cs:.0136,.576) {M};
\node[text=sbuBrick!85!black,anchor=north] at (axis cs:.0121,.530) {M$_Q$};
\end{scope}
\nextgroupplot[
  width=.52\textwidth,
  ybar=0pt, bar width=6.2pt,
  ymin=-.06, ymax=.6,
  ytick={0,.1,.2,.3,.4,.5},
  yticklabels={0,.1,.2,.3,.4,.5},
  ylabel={Fraction of the arm's gain},
  symbolic x coords={QA,QC,QD,QE,QB,IA,ID,IB},
  xtick=data,
  xticklabels={A,C,D,E,B,A,D,B},
  enlarge x limits=.07,
  ymajorgrids=true,
  legend style={draw=none,fill=white,fill opacity=.85,text opacity=1,at={(.99,.99)},anchor=north east,legend columns=1},
  title={\textbf{(b)} Attention rows of $C_k$: transplanted vs.\ restored},
  title style={font=\rmfamily\fontsize{8.4}{10}\selectfont,yshift=-2pt},
  clip=false,
]
\fill[sbuLine!25] ([xshift=-14pt]axis cs:QA,-.05) rectangle ([xshift=14pt]axis cs:IB,.08);
\addplot[draw=sbuTeal,fill=sbuTeal,line width=.3pt]
  coordinates {(QA,.391) (QC,.151) (QD,.345) (QE,.404) (QB,.041) (IA,.313) (ID,.284) (IB,.066)};
\addlegendentry{\textsc{Transplant} into base: recovered}
\addplot[draw=sbuBrick,fill=white,line width=.8pt]
  coordinates {(QA,.415) (QC,.109) (QD,.161) (QE,.171) (QB,.086) (IA,.053) (ID,.030) (IB,-.023)};
\addlegendentry{\textsc{Restore} in trained model: removed}
\draw[sbuSlate!70,line width=.4pt,densely dotted] ([xshift=14pt]axis cs:QB,-.06) -- ([xshift=14pt]axis cs:QB,.6);
\node[anchor=north,font=\rmfamily\fontsize{7.6}{9}\selectfont,text=sbuSlate] at ([yshift=-15pt]axis cs:QD,-.06) {\qwen{8B}};
\node[anchor=north,font=\rmfamily\fontsize{7.6}{9}\selectfont,text=sbuSlate] at ([yshift=-15pt]axis cs:ID,-.06) {\ivl{}};
\end{groupplot}
\end{tikzpicture}
\caption{\textbf{The repair runs through the comparison heads.} \textbf{(a)} Each point is one trained model (seed-averaged); selectivity is measured over the fixed heads $C_k$ as in \S\ref{sec:res_sel}. Open markers: A$_Q$ and E. Diamonds: attention-matched arms M and M$_Q$ (mean over three head sets), added later and therefore an out-of-sample check. Accuracy rises with selectivity in both families, including for the all-layer adapters D and E, which are not restricted to $C_k$. \textbf{(b)} Order-0 \spt{} test, seed~0. Filled bars: fraction of each arm's gain obtained by inserting only the image-2 attention rows of $C_k$ from the trained model into the base model. Open bars: fraction lost by restoring those rows to the base model inside the trained model. Grey band: range of both fractions over five layer-matched random head sets, across arms. As expected, arms whose training changed other heads (B) or only values (C) are hardly recovered by transplanting $C_k$ attention.}
\label{fig:repair}
\end{figure*}
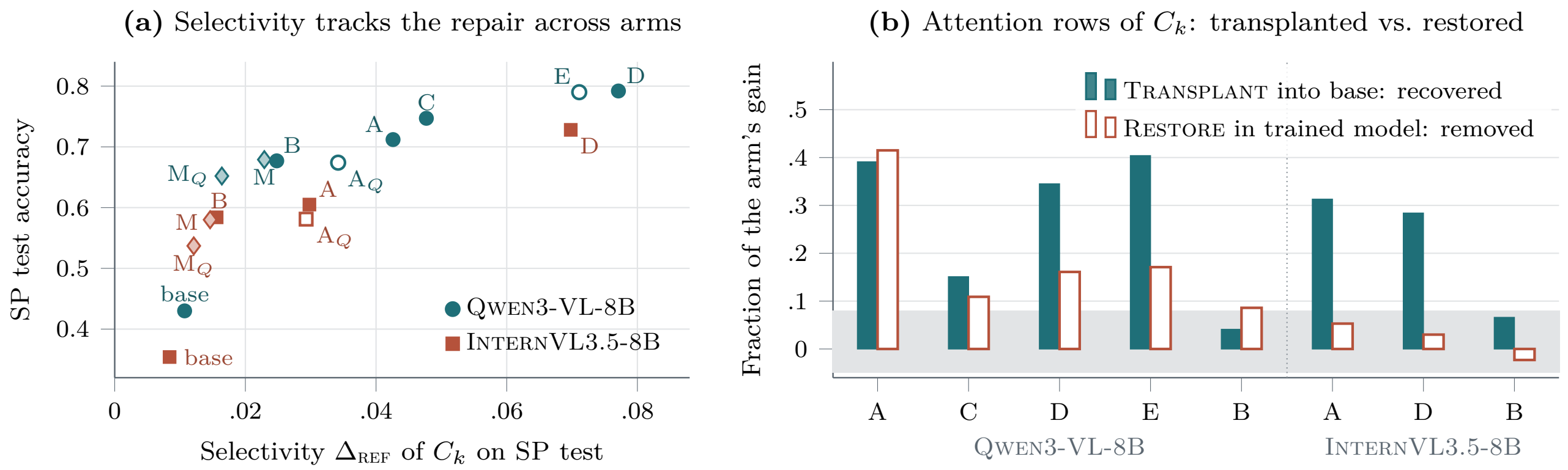

\subsection{The Repair Increases Selectivity}
\label{sec:mediation}

If the repair works by improving the comparison itself, the heads that carry it should become more selective, not merely more active. Arm A raises the selectivity of $C_k$ on \spt{} from $.0107$ to $.0426$ in \qwen{8B} and from $.0084$ to $.0298$ in \ivl{} (Table~\ref{tab:repaircmp}), more than twice the rise under the random-head arm B ($+.014$ and $+.007$) at similar accuracy. Across the base model and all trained arms, selectivity measured over the fixed heads $C_k$ orders \spt{} accuracy almost perfectly (Fig.~\ref{fig:repair}a; Spearman $0.96$ over seven \qwen{8B} models and $0.90$ over five \ivl{} models). The attention-matched arms M and M$_Q$ of \S\ref{sec:res_repair}, which were added later, give an out-of-sample check: including them, Spearman is $0.93$ over nine and $0.96$ over seven models. The relation includes D and E, which can change any layer; whatever else these adapters do, they make the comparison heads more selective. Training also raises the total attention that options send to image~1 (for the target, from $.199$ to $.437$ under A and $.535$ under D in \qwen{8B}), so the rise in selectivity is not merely a matter of scale: the ratio $\rref$, which total attention does not affect, rises from $1.30$ to $1.86$ (A) and $3.04$ (D) in \qwen{8B} and from $1.11$ to $1.53$ and $2.39$ in \ivl{}, and orders accuracy across all arms (Spearman $0.95$ over nine and $0.89$ over seven models; Table~\ref{tab:selmeasures}). The change comes mostly from the distractors. The share of a distractor's image-1 attention that falls on \refm{} drops from $.198$ to $.123$ (A) and $.072$ (D) in \qwen{8B}, while the target's share stays between $.21$ and $.24$; in \ivl{} both shares fall, the distractors' more ($.315$ to $.114$ and $.107$, against $.331$ to $.168$ and $.234$). Training thus largely stops the marked reference from drawing the wrong options, the salience effect of \S\ref{sec:res_sel}. Training the attention-matched heads M, which leaves the weights of $C_k$ untouched, also raises the selectivity of $C_k$ (to $.019$--$.029$ in \qwen{8B} and $.012$--$.017$ in \ivl{}), by less than A does. Because selectivity is read from the attention of a fixed head set, it needs neither a probe nor further training, which makes it a direct check of whether a candidate repair acts on the comparison step. Zeroing the A adapters on half of $C_k$ lowers \spt{} accuracy by $5.8$--$9.4\pp$ in \qwen{8B} and $4.2$--$6.1\pp$ in \ivl{} over three random halves, and by $6.0$ and $10.7\pp$ for the half with the highest retrieval score: the repair is distributed across the whole set rather than carried by a few heads.

\subsection{Does Unconstrained Training Converge on the Same Heads?}
\label{sec:convergence}

The targeted adapter is confined to these heads by design. The stronger test is whether an adapter with no such constraint ends up using them. We now test the relation in Fig.~\ref{fig:repair}a causally, by intervening on attention. For a trained arm with gain $g$ over the base model, we manipulate only the post-softmax attention rows of the heads in $C_k$ for image-2 query positions, leaving values, other heads, and all other positions untouched. \textsc{Transplant} runs the base model with these rows taken from the trained model on the same input; the \emph{recovered} fraction is its accuracy gain over base divided by $g$. \textsc{Restore} runs the trained model with these rows taken from the base model; the \emph{removed} fraction is its accuracy loss divided by $g$. Each operation is repeated on five layer-matched random head sets. We call a head set \emph{sufficient} (or \emph{necessary}) for an arm if the recovered (or removed) fraction is at least $0.25$, its numerator is positive under the bootstrap, and it exceeds all five random sets. The informative case is arm D, whose training is not restricted to $C_k$.

\paragraph{The comparison heads are sufficient.}
Transplanting the $C_k$ attention rows of the all-layer adapter D into the base model recovers $.345$ of its $+34.8\pp$ gain in \qwen{8B}, and $.466$ and $.514$ for two further training seeds; for \ivl{} it recovers $.284$, $.295$, and $.213$ over the three seeds (seed-0 gain $+37.0\pp$). The rank-16 adapter E gives $.404$ in \qwen{8B}. Random head sets recover between $-0.04$ and $+0.07$ (Fig.~\ref{fig:repair}b). In accuracy terms, for the arms that meet the sufficiency criterion and all their seeds, the base model with transplanted rows reaches $.549$--$.619$ on \spt{} (\qwen{8B}, base $.436$) and $.425$--$.463$ (\ivl{}, base $.354$), against $.42$--$.46$ and $.35$--$.37$ for random transplants. Transplanted states stay in distribution: $1$--$5\%$ of items fall outside the base residual-norm band, against $2\%$ for the base model itself. The effect extends to \hp{}, where transplanting the rows of D recovers $.612$ (\qwen{8B}) and $.481$ (\ivl{}) of D's gain, and those of E recover $.789$. For seed~0, the sufficiency criterion holds for A, D, and E in \qwen{8B} and for A and D in \ivl{}. It does not hold for B and C, whose training changed other heads or only values: what transplants is a change in the attention of these heads, not training as such.

\paragraph{The adapter changes the heads' inputs.}
An attention row depends on the head's query--key weights and on the residual stream it reads. We split \textsc{Transplant} accordingly. \textsc{Upstream} computes the rows of $C_k$ with the base model's query--key weights applied to the trained model's inputs. \textsc{Local} applies the trained weights to the base model's inputs. Each set of rows is then transplanted into the base model as before (Table~\ref{tab:decomp}). For D on \spt{}, \textsc{Upstream} recovers $.282$ of the gain in \qwen{8B} and $.233$ in \ivl{}, $83\%$ and $86\%$ of the full transplant measured in the same run, whereas \textsc{Local} recovers $.043$ and $.035$ with intervals reaching zero, and the interaction is near zero. E and the \hp{} items show the same ordering, and no random head set recovers more than $7.4\%$ of the gain on \spt{} under any operation. The all-layer adapter therefore acts on these heads mainly by changing their inputs. Their original weights, applied to the new inputs, already produce most of the transplanted effect.

\paragraph{Other routes compensate when the heads are reset.}
Restoring the base attention rows inside D removes $.161$ of its gain in \qwen{8B} ($.133$ and $.166$ for the other seeds) and $.022$--$.049$ in \ivl{}, below the necessity criterion in both models. For arm A, which changes only these heads' queries and the keys of their groups, restoring the image-2 rows removes $.347$--$.415$ in \qwen{8B} over three seeds but only $.053$ in \ivl{}. Restoring the heads' attention at \emph{all} query positions removes $.746$ and $.370$ of A's gain (seed~0; Table~\ref{tab:restoreall}), more than any of five random head sets (at most $.139$ and $.004$), so arm A meets the necessity criterion in both models (Appendix~\ref{app:edits} also restores every query head that shares their keys). For D, restoring all positions removes $.198$ and $.080$, more than any random set (at most $.063$ and $.037$) and below the criterion. The residual-norm check does not apply to \textsc{Restore}, because the trained models themselves lie outside the base band.

\paragraph{A convergent pathway.}
Among the things the unconstrained adapter learns is a change in how the comparison heads attend across images. In \qwen{8B}, transplanting this component alone recovers a third to a half of the gain, and it arises upstream of the heads rather than in their own query--key weights. The trained models also improve through additional routes, which compensate when the heads are reset. The targeted repair depends on the comparison heads; the unconstrained repair converges on them without depending on them alone. The heads are thus sufficient for a large part of the repair and form a convergent rather than an exclusive pathway.

\section{Discussion}
\label{sec:discussion}

\paragraph{Two failure points.}
For exact copies, the heads compare the two images well and the answer position carries the result, but the output does not use it, and neither calibration nor amplifying the decoded signal changes that. For semantic correspondence, the visual tokens hold the information, but the cross-image heads compare reference and candidates only weakly, and the result never reaches the answer position. Same-instance correspondence lies in between and is largely solved. For the semantic case, the repair experiments close the loop: changing only the similarity computed by the comparison heads recovers most of the deficit, raises their selectivity, and transfers to other correspondence benchmarks, and an adapter that is free to change anything still reshapes the attention of these heads in a way that reproduces a large share of its gain. Table~\ref{tab:claims} in Appendix~\ref{app:summary} lists the evidence and controls for each finding.

\paragraph{Relation to prior accounts.}
Our privileged readouts reproduce the large gaps reported with annotated-region readouts. On real correspondence, these gaps reflect information that the language model never brings to the answer, not an answer that is computed and then withheld, and this holds whether or not the entities can be named. The picture agrees with the observation that perceptual information inside the language model often goes unused \citep{fu2025hidden,liu2025visualrep}. What we add is the step at which this happens for cross-image tasks, and a component that can be trained to change it. The comparison heads resemble the gaze heads found for single-image description \citep{gandikota2026gaze} in that a small set of heads controls what reaches the answer. They differ in that they compare regions across images and form a pathway on which unconstrained fine-tuning converges.

\paragraph{Targeted and distributed repair.}
A query--key update changes which existing features the heads match. Sharpening the heads does not help, while learning their similarity does, so the comparison function is a limiting factor in its own right. An all-layer adapter can also change the features themselves and the path from the comparison to the answer, and it reaches higher accuracy by combining these routes. The decomposition in \S\ref{sec:convergence} shows how the two kinds of repair meet: the part of D's gain that passes through $C_k$ comes from changed inputs to the heads, which complements what an update of the heads' own projections can produce. Across every arm we trained, targeted or distributed, the selectivity of the same fixed heads tracked accuracy, which makes it a practical, training-free check on what fine-tuning has done to the comparison step.

\section{Conclusion}
\label{sec:conclusion}
We traced where cross-image correspondence is lost in VLMs. Exact copies are compared but the result is not read out, while semantic correspondence is present in the visual tokens and only weakly compared by a small set of cross-image heads. Training only these heads recovers most of the semantic deficit, and unconstrained adapters converge on the same heads, mostly by changing their inputs. Across every adapter we trained, the selectivity of this fixed head set tracked accuracy, giving a direct, training-free measure of what fine-tuning does to the comparison step.

\ifaclstyle
\section*{Limitations}
\else
\section*{Scope and Future Work}
\fi
Our mechanistic analyses and repairs focus on two model families at the 8B scale, with \gemma{} and the other \textsc{Qwen3-VL} sizes covering the measurement results. In our tasks the reference is made salient by a drawn marker, and the decoy- and moved-marker experiments characterize how this salience gates the comparison; references specified in language are a natural next setting. In the base models the semantic-correspondence behaviour runs mostly outside the comparison heads, and the trained models supply the causal evidence for that case. The answer-position experiments establish that the exact-copy comparison does not reach the output through the probe's direction; tracing the channel through which it is lost is a direct next step. Further directions include control head sets matched in layer and attention mass at once, general-capability evaluations after head-targeted training, and evaluation on newer correspondence benchmarks such as SOCO \citep{dunkel2026soco}.

\section*{Ethical Considerations}
This work analyses publicly released models on public or synthetic data. The adapters improve a perceptual capability on a narrow task family and introduce no new capability class. Head-targeted training changes model behaviour in ways that general evaluations may not reveal, as the synthetic-task drop under the targeted \ivl{} adapter shows; any such repair should be evaluated for unintended effects on unrelated behaviour before deployment.

\FloatBarrier
\ifaclstyle
\small
\else
\nobalance
\small
\setlength{\bibsep}{1.5pt plus .3pt minus .3pt}
\afterpage{\balance}
\fi

\clearpage
\nobalance
\appendix
\ifaclstyle\normalsize\else\onecolumn\fontsize{10}{12}\selectfont\fi

\Needspace{5\baselineskip}
\section{Summary of Findings}
\label{app:summary}

\begin{table}[H]
\centering
\caption{\textbf{Summary of findings, evidence, and controls.} Numbers are for \qwen{8B} / \ivl{} unless stated; pp are percentage points.}
\label{tab:claims}
\setlength{\tabcolsep}{4pt}
\begin{tabularx}{\textwidth}{P{4.1cm}YP{4.6cm}c}
\toprule
Finding & Evidence & Controls & Where \\
\midrule
Exact copies are compared but not read out & $\Pans-\Raff = +21.3\pp$ (\qwen{8B}, \syn{}-squiggle); $+15$ to $+25\pp$ in every family; answer-position edits move $\Acc$ by at most $0.2\pp$; probe-direction steering at the level of random directions & Output calibration; identical probe protocol on \hp{}/\spt{} (gap $\le 2.6\pp$); 20 magnitude-matched random subspaces; letter-direction positive control (switches $2.8$--$18.6\%$) & \S\ref{sec:res_ladder}--\ref{sec:res_breadth} \\
The gap is not about nameability & Gap difference squiggle vs.\ known: $+1.8\pp$ (\qwen{8B}), $+2.8\pp$ (\qwen{2B}) & Size- and stroke-matched nameable shapes & \S\ref{sec:res_ladder} \\
Semantic correspondence is present in visual tokens & $\Preg-\Pans$ on \spt{}: $+16.3$ / $+25.9\pp$, $+14.3\pp$ for \gemma{} & Marker-free images; layers chosen on development data & \S\ref{sec:res_priv} \\
A small set of cross-image heads is causal & Mass-preserving knockout of $C_k$: $\Acc$ $-17$ to $-22\pp$ on \syn{}, $-15$ to $-32\pp$ on \hp{} & 10 random and 3 attention-matched head sets ($\le 6.8\pp$); in-band items; within-image-2 knockout & \S\ref{sec:res_heads} \\
Their attention is salience-gated and content-weighted & $\dref$: \hp{} $.051$ / $.046$, \spt{} $.011$ / $.009$; decoy markers raise $\dref$ to $.109$; moved marker makes the new target selective ($.052$) & Held-out items; head sets fixed beforehand & \S\ref{sec:res_sel} \\
Training these heads repairs semantic correspondence & \spt{} $+28.3$ / $+25.1\pp$; transfer to \hp{} $+15.2$ / $+30.6\pp$ and BLINK & Layer-matched random heads (A$-$B $+3.6$ / $+2.1\pp$); attention-matched heads (A$-$M $+3.3$ / $+2.5\pp$); query-only arms (A$_Q-$B$_Q$ $+3.5$ / $+4.5\pp$, A$_Q-$M$_Q$ $+2.1$ / $+4.3\pp$) & \S\ref{sec:res_repair} \\
Transplanted head attention carries much of broad adapters' gain & Transplanting $C_k$ rows of D recovers $.345$--$.514$ / $.213$--$.295$ of its gain; $83$ / $86\%$ of it with base query--key weights & Five random head sets ($-0.04$ to $+0.07$); residual norms in distribution & \S\ref{sec:convergence} \\
The pathway is convergent, not exclusive & Restoring $C_k$ rows in D removes $.133$--$.166$ / $.022$--$.049$ & Five random head sets & \S\ref{sec:convergence} \\
\bottomrule
\end{tabularx}
\end{table}

\section{Notation}
\label{app:notation}

\begin{table}[H]
\centering
\caption{\textbf{Symbols used in the paper.}}
\label{tab:notation}
\begin{tabularx}{\textwidth}{P{3.3cm}Y}
\toprule
Symbol & Meaning \\
\midrule
$\Acc$ & Behavioural accuracy: argmax over the four letter log-probabilities at the scoring position. \\
$\Raff$ & Accuracy after multinomial logistic calibration of the four letter log-probabilities, fitted on development data. \\
$\Pans$ & Input-matched probe on the answer-position residual stream (PCA to 256 dimensions, multinomial logistic regression). \\
$\MaxSim$, $\Ppair$ & Privileged readouts over annotated regions: training-free MaxSim score and paired probe. \\
$\Preg$ & Region readout $\max(\MaxSim, \Ppair)$, with $\Ppair$ counted only if its selectivity over shuffled labels is at least $5\pp$. \\
$C_k$ & Cross-image head set: $C_{32}$ (32 heads, layers 1--24) for \qwen{8B}, $C_{16}$ (16 heads, layers 3--23) for \ivl{}. \\
$m_{\refm}(o)$ & Attention mass from the tokens of option $o$ to the \refm{} region, averaged over the heads in $C_k$. \\
$\dref$, $\rref$ & Selectivity: $m_{\refm}$ of the target minus the mean over distractors, and the ratio of the two means. \\
A, A$_Q$ & LoRA on the query--key (A) or query-only (A$_Q$) projections of $C_k$. \\
B, B$_Q$ & As A and A$_Q$, on three layer-matched random head sets of the same size. \\
M, M$_Q$ & As A and A$_Q$, on three head sets matched to $C_k$ in cross-image attention mass. \\
C & LoRA on the value projections of $C_k$. \\
D, E & LoRA on all attention projections of all layers, rank 2 (parameter-matched to A) and rank 16. \\
\textsc{Transplant}, \textsc{Restore} & Insert the trained model's image-2 attention rows of a head set into the base model, or the base model's rows into the trained model (\S\ref{sec:convergence}). \\
\textsc{Upstream}, \textsc{Local} & Transplanted rows computed with base query--key weights on trained inputs, or with trained weights on base inputs. \\
\textsc{Uniform}, \textsc{Shuffle}, \textsc{Block} & Knockouts of image-2$\to$image-1 attention: spread evenly or permuted (mass preserved), or removed. \\
\syn{}, \hp{}, \spt{} & Synthetic shape matching, HPatches point correspondence, SPair-71k semantic correspondence. \\
\bottomrule
\end{tabularx}
\end{table}

\section{Pre-Specified Criteria Not Met or Borderline}
\label{app:criteria}

Each analysis fixed its decision rules, thresholds, item sets, and statistical tests before it was run. Two amendments were made after a run had stopped: the trainable-parameter count of the attention-matched arms was reported instead of asserted equal to that of the reference arm (Appendix~\ref{app:controls}), and a reproduction tolerance for a different GPU type was relaxed, after which the attention maps were computed on the original GPU type at the original tolerance. Neither changed a reported result. Table~\ref{tab:criteria} lists every pre-specified criterion that was not met or was decided within a small margin, with what it means for the findings of Table~\ref{tab:claims}.

\begin{table}[H]
\centering
\caption{\textbf{Pre-specified criteria not met or borderline.}}
\label{tab:criteria}
\begin{tabularx}{\textwidth}{P{2.9cm}YY}
\toprule
Analysis & Criterion and outcome & Bearing on the findings \\
\midrule
Readout classification & A frozen rule assigns the \syn{} gap either to the readout (the gap survives calibration) or to retention and computation; \qwen{2B} falls within $0.2\pp$ of the boundary. & The readout account rests on \qwen{8B}, which is far from the boundary (Table~\ref{tab:calib}). \\
Region readout on \spt{} & Presence threshold: region readout at least $15\pp$ above the answer-position probe. Met by \qwen{8B} ($+16.3$) and \ivl{} ($+25.9$); \gemma{} reaches $+14.3$. & The direction holds in all three models; the margin is smaller for \gemma{}. \\
Cross-image heads & Per-option retrieval: each option's attention should peak on its own correspondence, also on \spt{}, and this should predict correctness. Not met: options attend mostly to \refm{}. & This is the observation that led to the salience-gating account of \S\ref{sec:res_sel}. \\
Held-out confirmation & Switch: after the \refm{} marker is moved, the old target's selectivity should be non-positive; a residual of $+0.009$ remains. Decoys: the target's attention peak should stay on \refm{} when decoy markers are drawn; this did not hold in either model. Sharpening the heads ($\beta$) should help; it does not. & The residual is the content signal and the decoy shift is the salience effect, both reported in \S\ref{sec:res_sel}; the sharpening result motivates training the similarity rather than its temperature. \\
Targeted repair & A$-$B should exceed $3\pp$: \ivl{} reaches $+2.1\pp$ (interval excludes zero). A does not match the parameter-matched all-layer adapter D (also at 500 items); the value arm C exceeds A in \qwen{8B}; A lowers both \syn{} conditions in \ivl{}. & Localization holds in both models, with a smaller margin in \ivl{}; the targeted adapter is presented as a sufficient lever, not as the strongest adapter (\S\ref{sec:res_repair}). \\
Convergence & Necessity: restoring the $C_k$ rows should remove at least $0.25$ of the gain. Not met for D in either model; met for A only in \qwen{8B} (image-2 rows) and in both models at all positions. \ivl{} D seed~2 recovers $.213$ under transplant, below $0.25$. & The heads form a sufficient, convergent rather than exclusive pathway (\S\ref{sec:convergence}). \\
Answer-position edits & The underweighted-readout and used-on-\hp{} criteria were not met; the unused-on-\syn{} criterion was met. & The decoded signal is off the output path; on \hp{} the model answers through other channels (\S\ref{sec:res_breadth}). \\
Attention-matched controls & Strong localization: both A$-$M and A$_Q-$M$_Q$ at least $3\pp$. Not met in one comparison per model (A$_Q-$M$_Q$ $+2.1\pp$ in \qwen{8B}, A$-$M $+2.5\pp$ in \ivl{}). & All four differences are positive with intervals above zero. \\
Positive controls & The probe direction should steer the output beyond random directions, more on \hp{} than on \syn{}. Not met. & This is the finding of \S\ref{sec:res_breadth}: the probe's direction is not the one the output reads. \\
Mass-preserving knockouts & In distribution: at most $5\%$ of items outside the norm band. Failed on \hp{} for every operation ($5.6$--$20.6\%$) and on \syn{} in \qwen{8B} under \textsc{Shuffle} ($5.6\%$) and \textsc{Block} ($49\%$). On \spt{}, the causal criterion held only for \textsc{Shuffle} and \textsc{Block} in \qwen{8B}. & Drops on in-range items are within $2\pp$ of the overall drops; the base models' semantic answers barely use the heads, which is the gap the repair closes (\S\ref{sec:res_heads}). \\
\bottomrule
\end{tabularx}
\end{table}

\section{Repair Details}
\label{app:repair}

\paragraph{Adapters.}
In arm A the query adapter of each layer acts only on the output dimensions of the selected heads in that layer; the key adapter acts on the key--value group of each selected head, so with grouped-query attention it also changes the keys seen by the other query heads of that group. Arm C applies the same restriction to the value projection. Arms D and E adapt the query, key, value, and output projections of every decoder layer. We confirmed that training changes only the intended adapter parameters and leaves the base weights unchanged.

\paragraph{Selection and data.}
Training pairs come from the SPair-71k training split and validation pairs from its validation split; test pairs are from the test split and were not used in any earlier run. The learning rate was chosen per arm on the validation items. For \qwen{8B} arm E, the largest learning rate diverged and was not selected. Accuracies in Table~\ref{tab:repair} average three seeds for A, C, and D and three random head sets for B; differences use a paired item bootstrap with seeds averaged first; hierarchical intervals resample the three random head sets and the items jointly. B$_Q$ uses the A$_Q$ recipe on the three B head sets, with the learning rate chosen on set~0 ($3\cdot10^{-4}$ for \qwen{8B}, $10^{-3}$ for \ivl{}).

\paragraph{Comparisons.}
Table~\ref{tab:repaircmp} lists the planned comparisons on the \spt{} test set.

\begin{table}[H]
\centering
\caption{\textbf{Planned comparisons on \spt{} test} (pp, paired 95\% intervals where available).}
\label{tab:repaircmp}
\begin{tabular}{lcc}
\toprule
Comparison & \qwen{8B} & \ivl{} \\
\midrule
A $-$ base ($N{=}2{,}000$) & $+28.3$ [$25.3$, $31.1$] & $+25.1$ [$22.4$, $27.8$] \\
A $-$ mean(B) & $+3.6$ [$2.0$, $5.2$] & $+2.1$ [$0.8$, $3.4$] \\
A $-$ mean(B), hierarchical & $+3.6$ [$1.4$, $6.0$] & $+2.1$ [$0.2$, $4.1$] \\
A$_Q$ $-$ mean(B$_Q$), hierarchical & $+3.5$ [$1.6$, $5.5$] & $+4.5$ [$2.1$, $7.0$] \\
A $-$ D & $-8.0$ & $-12.3$ \\
A $-$ C & $-3.5$ & -- \\
A $-$ E & $-7.8$ & -- \\
A $-$ base ($N{=}500$) & $+23.9$ & $+19.0$ \\
A $-$ D ($N{=}500$) & $-5.0$ & $-12.7$ \\
Selectivity rise, A & $+.032$ [$.029$, $.035$] & $+.021$ [$.016$, $.027$] \\
Selectivity rise, mean(B) & $+.014$ & $+.007$ \\
\bottomrule
\end{tabular}
\end{table}

\begin{table}[H]
\centering
\caption{\textbf{Selectivity of $C_k$ under three measures} (\spt{} test, 400 items, order 0; seeds or head sets averaged). $\dref$: target minus mean distractor attention to \refm{}; $\rref$: ratio of the two; share: fraction of an option's image-1 attention that falls on \refm{}; image-1 mass: the target's total attention to image~1.}
\label{tab:selmeasures}
\setlength{\tabcolsep}{5pt}
\begin{tabular}{llcccccc}
\toprule
Model & Arm & \spt{} accuracy & $\dref$ & $\rref$ & Target share & Distractor share & Image-1 mass \\
\midrule
\qwen{8B} & base & .430 & .0107 & 1.297 & .238 & .198 & .199 \\
 & A & .712 & .0426 & 1.861 & .209 & .123 & .437 \\
 & A$_Q$ & .674 & .0342 & 1.623 & .225 & .156 & .396 \\
 & B & .677 & .0248 & 1.550 & .225 & .168 & .308 \\
 & C & .747 & .0477 & 2.200 & .205 & .106 & .420 \\
 & D & .792 & .0771 & 3.042 & .208 & .072 & .535 \\
 & E & .790 & .0711 & 3.017 & .200 & .072 & .513 \\
 & M & .679 & .0229 & 1.590 & .238 & .178 & .256 \\
 & M$_Q$ & .652 & .0164 & 1.457 & .221 & .170 & .237 \\
\midrule
\ivl{} & base & .354 & .0084 & 1.112 & .331 & .315 & .248 \\
 & A & .605 & .0298 & 1.526 & .168 & .114 & .517 \\
 & A$_Q$ & .581 & .0293 & 1.345 & .218 & .170 & .524 \\
 & B & .584 & .0156 & 1.225 & .298 & .266 & .282 \\
 & D & .728 & .0698 & 2.387 & .234 & .107 & .500 \\
 & M & .580 & .0146 & 1.229 & .299 & .269 & .269 \\
 & M$_Q$ & .537 & .0121 & 1.166 & .324 & .301 & .259 \\
\bottomrule
\end{tabular}
\end{table}

\section{Attention Transplantation: Full Results}
\label{app:convergence}

Table~\ref{tab:conv} gives the restore and transplant fractions for every analysed arm and seed on the 1{,}000 \spt{} test items (order 0). Random head sets give fractions between $-0.05$ and $+0.08$ for every arm.

\begin{table}[H]
\centering
\caption{\textbf{Removed and recovered fractions for $C_k$}, \spt{} test, order 0.}
\label{tab:conv}
\begin{tabular}{llcccc}
\toprule
Model & Arm & Seed & Gain (pp) & Removed by \textsc{Restore} & Recovered by \textsc{Transplant} \\
\midrule
\multirow{9}{*}{\qwen{8B}}
 & A & 0 & $+28.9$ & $.415$ [$.333$, $.504$] & $.391$ \\
 & A & 1 & & $.347$ & $.434$ \\
 & A & 2 & & $.356$ & $.434$ \\
 & C & 0 & $+31.1$ & $.109$ & $.151$ \\
 & D & 0 & $+34.8$ & $.161$ [$.101$, $.220$] & $.345$ [$.273$, $.413$] \\
 & D & 1 & & $.133$ & $.466$ \\
 & D & 2 & & $.166$ & $.514$ \\
 & E & 0 & $+35.6$ & $.171$ & $.404$ \\
 & B & set 0 & $+24.3$ & $.086$ & $.041$ \\
\midrule
\multirow{5}{*}{\ivl{}}
 & A & 0 & $+22.7$ & $.053$ & $.313$ \\
 & D & 0 & $+37.0$ & $.030$ & $.284$ \\
 & D & 1 & & $.022$ & $.295$ \\
 & D & 2 & & $.049$ & $.213$ \\
 & B & set 0 & -- & $-.023$ & $.066$ \\
\bottomrule
\end{tabular}
\end{table}

\section{Answer-Position Edits and Full-Position Restore}
\label{app:edits}

\paragraph{Answer-position edits.}
For each model and task, the answer-position probe of \S\ref{sec:res_ladder} was refitted from the stored development features and reproduces the original probe's per-item predictions exactly. $S$ is the span of its centred class directions mapped back to the residual space (rank~3). Edits act on the answer-position residual at the probe's layer (\qwen{8B}: 31 for \syn{}, 28 for \hp{}; \ivl{}: 32 and 33): \textsc{Ablate} sets $h\leftarrow h-P_S(h-\mu)$ and \textsc{Amplify} sets $h\leftarrow h+(\gamma-1)P_S(h-\mu)$, with $\mu$ the development mean. Each of 20 random rank-3 subspaces from the top-256 principal components is scaled so that its edit moves the residual by the same mean squared distance as $S$ on development data. Table~\ref{tab:edits} reports the change in $\Acc$; calibrated accuracy changes by at most $0.2\pp$ in every cell. Between 1.7\% and 2.4\% of items edited along $S$ leave the base residual-norm band (1.2\% to 4.2\% for the random subspaces), against 2.0\% by construction for unedited items.

\begin{table}[H]
\centering
\caption{\textbf{Answer-position edits along the probe subspace} (test, $n=1{,}000$, order 0). $\Delta\Acc$ in pp; random range over 20 subspaces and the three operations.}
\label{tab:edits}
\begin{tabular}{llccccc}
\toprule
Model & Task & Base $\Acc$ & \textsc{Ablate} & \textsc{Amplify} $\gamma{=}2$ & \textsc{Amplify} $\gamma{=}4$ & Random range \\
\midrule
\qwen{8B} & \syn{}-squiggle & .413 & $+0.0$ & $+0.0$ & $-0.1$ & $-0.2$ to $+0.1$ \\
\qwen{8B} & \hp{} & .767 & $-0.2$ & $+0.1$ & $+0.1$ & $-0.3$ to $+0.3$ \\
\ivl{} & \syn{}-squiggle & .514 & $+0.0$ & $+0.0$ & $+0.1$ & $-0.2$ to $+0.3$ \\
\ivl{} & \hp{} & .543 & $-0.1$ & $+0.0$ & $+0.0$ & $-0.5$ to $+0.4$ \\
\bottomrule
\end{tabular}
\end{table}

\paragraph{Full-position restore.}
\textsc{Restore-all} replaces the attention of the heads in $C_k$ at every query position with that of the base model; \textsc{Restore-group} does so for every query head in the key--value groups that $C_k$ touches. Table~\ref{tab:restoreall} gives the removed fractions on the 1{,}000 \spt{} test items (order 0, seed 0) against the five random head sets of \S\ref{sec:convergence}. For arm A, \textsc{Restore-group} reproduces the base model's log-probabilities on every item in both models. Random sets share some key--value groups with $C_k$ (four to nine of 20 in \qwen{8B}, at most two of 14 in \ivl{}), so random \textsc{Restore-group} values are not a clean control (in \qwen{8B} they range from .125 to .606 for arm A, without a monotone relation to the overlap) and enter no criterion. The gains in Table~\ref{tab:restoreall} differ from the seed-0 gains in Table~\ref{tab:conv} by at most $0.4\pp$, because a larger batch size changes bf16 rounding on a few items.

\begin{table}[H]
\centering
\caption{\textbf{Fraction of the gain removed by full-position restore} (\spt{} test, order 0, seed 0; 95\% intervals).}
\label{tab:restoreall}
\setlength{\tabcolsep}{4pt}
\begin{tabular}{llcccc}
\toprule
Model & Arm & Gain (pp) & \textsc{Restore-all}$(C_k)$ & Random \textsc{Restore-all} & \textsc{Restore-group}$(C_k)$ \\
\midrule
\qwen{8B} & A & $+28.7$ & $.746$ [$.671$, $.823$] & $.024$ to $.139$ & $1.000$ \\
\qwen{8B} & D & $+34.9$ & $.198$ [$.137$, $.260$] & $-.006$ to $.063$ & $.183$ [$.122$, $.245$] \\
\ivl{} & A & $+22.7$ & $.370$ [$.248$, $.479$] & $-.013$ to $.004$ & $1.000$ \\
\ivl{} & D & $+37.4$ & $.080$ [$.029$, $.135$] & $-.013$ to $.037$ & $.120$ [$.064$, $.172$] \\
\bottomrule
\end{tabular}
\end{table}

\section{Attention Maps of Repaired Models}
\label{app:maps}

\begin{figure}[H]
\centering
\includegraphics[width=\textwidth]{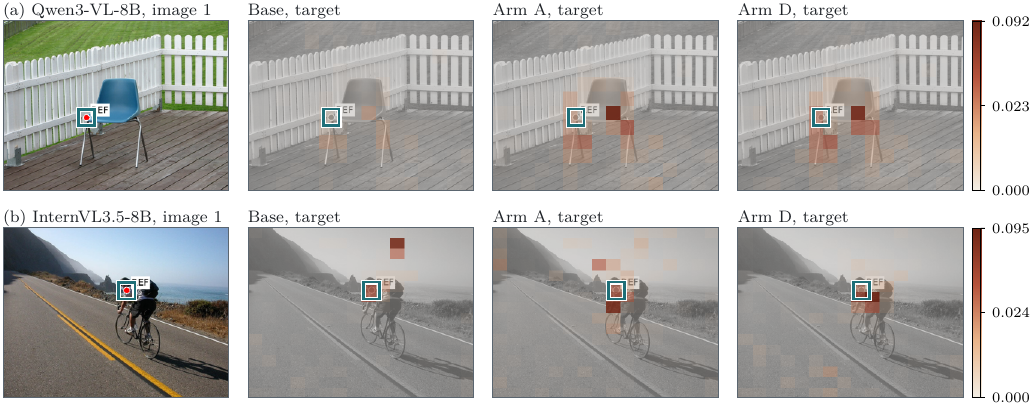}
\caption{\textbf{Target-option attention before and after repair} on one \spt{} repair item per model (mean over $C_k$; square-root colour scale shared within a row; teal: \refm{}).}
\label{fig:qualrepair}
\end{figure}

Figure~\ref{fig:qualrepair} shows the target option's attention for one \spt{} repair item per model, before and after training. The item was chosen before any map was computed: among the first 400 repair test pairs, the one whose change in $\dref$ under arm A (seed~0) is closest to the median change. All three models answer it correctly. In \qwen{8B}, the target's attention to \refm{} rises from $.011$ (base) to $.048$ (A) and $.072$ (D), while its total attention to image~1 rises from $.16$ to $.51$ and $.56$; the single most attended token after training lies two tokens to the right of \refm{}, on the location corresponding to distractor D, whose attention mass rises from $.024$ to $.119$ (A) and $.116$ (D). On this item the share of image-1 attention on \refm{} therefore rises less (from $7\%$ to $9.5\%$ and $13\%$) than the absolute mass, and the map shows that training raises attention to image~1 broadly as well as to \refm{}. Over all 400 items, the scale-free measures of \S\ref{sec:mediation} separate the two effects: the ratio $\rref$ rises in every trained arm, and the distractors' share of image-1 attention on \refm{} falls (Table~\ref{tab:selmeasures}). In \ivl{}, the attention to \refm{} rises from $.120$ to $.141$ and $.201$ and concentrates on the \refm{} region under D. Maps for six randomly drawn items per task and model, for the \ivl{} counterpart of Fig.~\ref{fig:qual}, and for the decoy condition on \spt{} will be released with the code.

\section{Additional Controls and Analyses}
\label{app:controls}

This appendix gives details for four analyses: attention-matched control heads, positive controls at the answer position, the decomposition of \textsc{Transplant}, and the mass-preserving knockouts. The attention-matched head sets are drawn from any layer rather than from the layers of $C_k$, and their trainable-parameter counts are reported rather than equalized.

\paragraph{Attention-matched heads.}
For each head $h$, $X(h)$ is the attention mass from image-2 query positions to all image-1 key positions, averaged over \syn{} development items; $R(h)$ is the correspondence score used to select $C_k$. The three sets $M_0$--$M_2$ are disjoint and built greedily: for each head $c\in C_k$ in turn, the head with $X$ closest to $X(c)$ among those not in $C_k$, not among the $3|C_k|$ heads with the highest $R$, and not yet used. Table~\ref{tab:matchedsets} compares them with the layer-matched random sets of arm B. Because the matched heads are spread over more key--value groups, every matched arm has more trainable parameters than its reference (\qwen{8B}: M $1.96$--$2.75$M vs.\ $1.68$M, M$_Q$ $0.98$--$1.38$M vs.\ $0.85$M; \ivl{}: $1.64$--$1.77$M vs.\ $1.50$M, $0.82$--$0.88$M vs.\ $0.75$M). Training follows the recipe of \S\ref{sec:res_repair} with one seed per set; the learning rate was chosen on set~0 from the same grid ($3\cdot10^{-4}$ for \qwen{8B} M, $10^{-3}$ for the other three arms). The grid's largest rate was thus selected in three of four sweeps, as for A in \ivl{}; A$_Q$ reuses A's rate in both models, without a sweep of its own. Table~\ref{tab:matched} gives the comparisons. In \qwen{8B}, M$_Q$ closes about $40\%$ of the gap between A$_Q$ and B$_Q$; in the other three comparisons matched heads do about as well as random ones. On the synthetic conditions, which no arm is trained on, \ivl{} M and M$_Q$ are $10.7$--$16.3\pp$ above A and A$_Q$, extending the collateral drop of the targeted \ivl{} arms noted in \S\ref{sec:res_repair}; on \hp{} and BLINK-SC they are below.

\begin{table}[H]
\centering
\caption{\textbf{Head-set statistics relative to $C_k$.} Ratios of set means; $X_{\spt{}}$ is measured on 200 \spt{} development items.}
\label{tab:matchedsets}
\begin{tabular}{llccc}
\toprule
Model & Set & $X$ ratio & $R$ ratio & $X_{\spt{}}$ ratio \\
\midrule
\qwen{8B} & M$_0$ / M$_1$ / M$_2$ & .985 / .950 / .803 & .107 / .043 / .081 & 1.670 / 1.575 / 1.319 \\
 & B$_0$ / B$_1$ / B$_2$ & .350 / .430 / .368 & .131 / .152 / .127 & .448 / .554 / .487 \\
\ivl{} & M$_0$ / M$_1$ / M$_2$ & .969 / .875 / .827 & .101 / .125 / .105 & 1.709 / 1.390 / 1.401 \\
 & B$_0$ / B$_1$ / B$_2$ & .316 / .467 / .464 & .017 / .115 / .112 & .522 / .701 / .717 \\
\bottomrule
\end{tabular}
\end{table}

\begin{table}[H]
\centering
\caption{\textbf{$C_k$ against attention-matched heads} (\spt{} test, 1{,}000 items, mean over four letter orders, \%). Paired intervals average the three sets first; hierarchical intervals resample sets and items.}
\label{tab:matched}
\small\setlength{\tabcolsep}{3pt}
\begin{tabular}{llcccccc}
\toprule
Model & Comparison & $C_k$ arm & Matched (per set) & B mean & Difference & Paired CI & Hierarchical CI \\
\midrule
\qwen{8B} & A $-$ M & 71.24 & 67.90 (68.28, 67.40, 68.03) & 67.66 & $+3.34$ & [$1.73$, $4.94$] & [$1.62$, $5.07$] \\
 & A$_Q$ $-$ M$_Q$ & 67.35 & 65.23 (65.03, 66.03, 64.65) & 63.82 & $+2.12$ & [$0.50$, $3.78$] & [$0.30$, $4.02$] \\
\ivl{} & A $-$ M & 60.51 & 58.03 (57.38, 59.18, 57.53) & 58.38 & $+2.48$ & [$1.19$, $3.84$] & [$0.75$, $4.24$] \\
 & A$_Q$ $-$ M$_Q$ & 58.05 & 53.73 (52.78, 54.87, 53.52) & 53.52 & $+4.32$ & [$2.77$, $5.98$] & [$2.42$, $6.39$] \\
\bottomrule
\end{tabular}
\end{table}

\begin{table}[H]
\centering
\caption{\textbf{Transfer and collateral accuracy of the matched arms} (\%, mean over the three sets; A and A$_Q$ from Table~\ref{tab:repair}).}
\label{tab:matchedtransfer}
\begin{tabular}{llccccc}
\toprule
Model & Arms & \hp{} & BLINK-SC & BLINK-VC & \syn{}-squiggle & \syn{}-known \\
\midrule
\qwen{8B} & A / M & 93.4 / 91.7 & 54.9 / 48.9 & 80.2 / 79.8 & 55.1 / 51.1 & 83.3 / 79.1 \\
 & A$_Q$ / M$_Q$ & 92.2 / 89.7 & 50.4 / 45.6 & 79.1 / 74.6 & 48.0 / 49.5 & 84.2 / 79.0 \\
\ivl{} & A / M & 83.7 / 82.5 & 36.7 / 33.6 & 56.2 / 57.2 & 36.6 / 47.3 & 77.6 / 91.2 \\
 & A$_Q$ / M$_Q$ & 83.7 / 79.0 & 41.0 / 34.8 & 55.8 / 52.9 & 31.5 / 47.8 & 78.4 / 92.9 \\
\bottomrule
\end{tabular}
\end{table}

\paragraph{Answer-position positive controls.}
On the first 500 test items of \syn{}-squiggle and \hp{} (order 0), at the position and layer of Appendix~\ref{app:edits}, edits add $\alpha s\,u$ to the residual, with $s$ the standard deviation of the residual within the probe subspace $S$, $\alpha\in\{1,2,4,8\}$, and $u$ a unit direction. For the letter direction, $u\propto W_U[c]-\overline{W_U[c']}_{c'\neq c}$, where $c$ is the first letter that is not the model's base answer; the switch rate is the fraction of items whose answer becomes $c$. For the probe direction, $u$ is the class direction of the probe for the first class it does not predict, mapped back to the residual space; the steer rate is the fraction of items whose answer becomes that class's letter. Because this rate also counts items already answered with that letter ($4.4$--$10.0\%$, identical in every row), Table~\ref{tab:posctrl} also reports the rate on the remaining items. Controls are ten random unit directions from the top-256 principal components at the same $\alpha s$. At most $4.8\%$ of letter- or probe-edited items leave the base norm band ($10.2\%$ for random directions at $\alpha=8$). At no $\alpha$, in either model, does the probe direction exceed the random mean by $5\pp$ or more, and the \hp{}$-$\syn{} difference in steer rate has an interval including zero ($-1.2\pp$ [$-4.0$, $1.4$] and $+2.2\pp$ [$-1.4$, $5.8$]).

\begin{table}[H]
\centering
\caption{\textbf{Answer-position positive controls} (\%, 500 test items per task). Last three columns at $\alpha=8$. Moved: steer rate on the items not already answered with the target letter.}
\label{tab:posctrl}
\small\setlength{\tabcolsep}{3pt}
\begin{tabular}{llcccc}
\toprule
Model & Task & Letter switch, $\alpha=1/2/4/8$ & Random max & Probe moved & Random moved \\
\midrule
\qwen{8B} & \syn{}-squiggle & 0.4 / 0.4 / 0.8 / 2.8 & 0.4 & 0.4 & 0.2--0.4 \\
 & \hp{} & 0.8 / 1.2 / 3.0 / 9.6 & 0.8 & 0.4 & 0.0--0.6 \\
\ivl{} & \syn{}-squiggle & 0.6 / 0.6 / 1.8 / 3.8 & 0.4 & 0.9 & 0.0--1.5 \\
 & \hp{} & 3.2 / 4.6 / 9.4 / 18.6 & 1.8 & 1.6 & 0.9--1.6 \\
\bottomrule
\end{tabular}
\end{table}

\paragraph{Decomposition of \textsc{Transplant}.}
\textsc{Upstream} runs the trained model and computes the image-2 rows of $C_k$ with the query--key adapter updates switched off for these rows only; \textsc{Local} runs the base model and computes them with the updates switched on. Rows are transplanted into the base model as in \S\ref{sec:convergence}, on the 1{,}000 \spt{} and 400 \hp{} items of that section (order 0, seed 0), and the same three operations are applied to the five random head sets. Fractions use numerator and gain from the same bootstrap resamples. These runs use a larger batch size, which changes bf16 rounding on 1--19 items per arm, so \textsc{Full} differs slightly from Table~\ref{tab:conv}. For D on \hp{} in \qwen{8B}, \textsc{Upstream} exceeds \textsc{Full} and the interaction interval excludes zero.

\begin{table}[H]
\centering
\caption{\textbf{Recovered fraction of each arm's gain} under transplanted $C_k$ rows (95\% intervals). Interaction $=$ \textsc{Full} $-$ \textsc{Upstream} $-$ \textsc{Local}. Rank: position of $C_k$ among $C_k$ and five random sets (\textsc{Full} / \textsc{Upstream} / \textsc{Local}).}
\label{tab:decomp}
\footnotesize\setlength{\tabcolsep}{2.5pt}
\begin{tabular}{lllcccccc}
\toprule
Model & Arm & Split & Gain (pp) & \textsc{Full} & \textsc{Upstream} & \textsc{Local} & Interaction & Rank \\
\midrule
\qwen{8B} & D & \spt{} & 34.8 & .339 [.268, .404] & .282 [.207, .348] & .043 [$-$.013, .095] & $+$.014 [$-$.059, .088] & 1 / 1 / 1 \\
 & D & \hp{} & 16.2 & .585 [.417, .759] & .692 [.550, .857] & .077 [$-$.062, .207] & $-$.185 [$-$.339, $-$.034] & 1 / 1 / 1 \\
 & E & \spt{} & 35.3 & .411 [.340, .484] & .360 [.291, .435] & .028 [$-$.012, .066] & $+$.023 [$-$.031, .078] & 1 / 1 / 2 \\
 & E & \hp{} & 16.8 & .806 [.672, .953] & .791 [.646, .962] & .090 [.000, .190] & $-$.075 [$-$.213, .061] & 1 / 1 / 2 \\
\ivl{} & D & \spt{} & 37.4 & .270 [.201, .339] & .233 [.166, .298] & .035 [.000, .068] & $+$.003 [$-$.041, .054] & 1 / 1 / 1 \\
 & D & \hp{} & 39.0 & .462 [.366, .558] & .442 [.344, .539] & .026 [$-$.029, .077] & $-$.006 [$-$.078, .069] & 1 / 1 / 1 \\
\bottomrule
\end{tabular}
\end{table}

\paragraph{Mass-preserving knockouts.}
Operations act on the post-softmax image-2 query rows of a head set, image-1 key block only. \textsc{Uniform} replaces the distribution over image-1 keys by a uniform one with the same total mass; \textsc{Shuffle} permutes the values over image-1 keys with one fixed permutation per item and layer; \textsc{Block} is the knockout of \S\ref{sec:res_heads}. Items are the first 500 test items per task (order 0); controls are the first ten layer-matched random sets of \S\ref{sec:res_heads} and the three attention-matched sets. Calibrated accuracy follows $\Acc$ within $4\pp$ in every cell. Under every operation, random sets leave the norm band on at most $5.6\%$ of items and matched sets on at most $8.0\%$.

\begin{table}[H]
\centering
\caption{\textbf{Knockouts of $C_k$, full results} (pp; paired 95\% intervals). Random, Matched: largest $\Acc$ drop among ten random and three attention-matched sets. Out: \% of items outside the base residual-norm band.}
\label{tab:knockfull}
\setlength{\tabcolsep}{4pt}
\begin{tabular}{llccccc}
\toprule
Task & Operation & $\Acc$ drop & $\Pans$ drop & Random & Matched & Out \\
\midrule
\multicolumn{7}{l}{\emph{\qwen{8B} (base $\Acc$ / $\Pans$: \syn{} 41.2 / 62.2, \hp{} 76.2 / 81.2, \spt{} 46.6 / 42.0)}} \\
\syn{} & \textsc{Uniform} & 17.2 [12.6, 22.0] & 38.2 [32.6, 43.8] & 2.8 & 0.6 & 4.6 \\
 & \textsc{Shuffle} & 18.2 [13.8, 22.8] & 37.6 [31.6, 43.2] & 3.0 & 0.2 & 5.6 \\
 & \textsc{Block} & 17.4 [12.2, 22.8] & 34.8 [28.8, 40.4] & 3.6 & $-$0.8 & 49.0 \\
\hp{} & \textsc{Uniform} & 32.2 [27.6, 37.0] & 35.0 [30.0, 39.8] & 1.2 & 1.2 & 5.6 \\
 & \textsc{Shuffle} & 31.8 [27.2, 36.4] & 32.4 [27.8, 37.6] & 1.0 & 1.8 & 5.8 \\
 & \textsc{Block} & 21.2 [17.4, 25.4] & 26.2 [21.6, 31.0] & 2.4 & 0.2 & 15.8 \\
\spt{} & \textsc{Uniform} & 2.2 [$-$0.4, 4.8] & 2.2 [$-$1.4, 6.0] & 1.0 & 0.0 & 1.8 \\
 & \textsc{Shuffle} & 3.2 [0.4, 6.0] & 2.2 [$-$1.2, 6.0] & 1.2 & $-$0.6 & 1.6 \\
 & \textsc{Block} & 3.6 [0.8, 6.6] & 4.2 [0.2, 8.4] & 1.4 & 2.0 & 1.0 \\
\midrule
\multicolumn{7}{l}{\emph{\ivl{} (base $\Acc$ / $\Pans$: \syn{} 48.6 / 73.4, \hp{} 53.6 / 54.0, \spt{} 34.8 / 29.4)}} \\
\syn{} & \textsc{Uniform} & 20.2 [15.4, 25.0] & 48.4 [43.6, 53.4] & 3.4 & 6.6 & 3.6 \\
 & \textsc{Shuffle} & 22.0 [17.2, 26.8] & 46.0 [41.0, 51.0] & 1.8 & 6.2 & 2.0 \\
 & \textsc{Block} & 22.4 [17.4, 27.2] & 46.4 [41.4, 51.4] & 2.6 & 6.8 & 2.0 \\
\hp{} & \textsc{Uniform} & 16.8 [13.0, 20.8] & 20.6 [16.2, 25.2] & 2.8 & 1.8 & 15.8 \\
 & \textsc{Shuffle} & 14.6 [10.8, 18.4] & 19.6 [15.0, 24.0] & 4.2 & 1.4 & 16.2 \\
 & \textsc{Block} & 16.2 [12.2, 20.6] & 18.2 [13.4, 23.0] & 2.4 & 1.8 & 20.6 \\
\spt{} & \textsc{Uniform} & $-$0.8 [$-$3.0, 1.6] & $-$1.2 [$-$3.8, 1.4] & 0.0 & 0.0 & 1.8 \\
 & \textsc{Shuffle} & $-$0.6 [$-$3.0, 1.8] & $-$1.6 [$-$4.2, 1.0] & 0.4 & 1.0 & 1.4 \\
 & \textsc{Block} & $-$0.6 [$-$3.0, 2.0] & $-$0.8 [$-$3.4, 1.8] & 0.2 & $-$0.2 & 1.4 \\
\bottomrule
\end{tabular}
\end{table}

\section{Reproducibility Details}
\label{app:repro}

Thresholds and decision criteria for each analysis were fixed before it was run. Every number in the main text was recomputed from the raw per-item outputs by analysis code kept separate from the experiment code; for probe and calibration refits, we checked that the refit reproduces the stored per-item predictions. Region-to-token mappings were validated with a perturbation test at the patch-embedding output; for InternVL3.5, the pixel footprint of each region was defined from the support of the resize kernel. The held-out items of \S\ref{sec:res_sel} use seeds, sequences, and image pairs disjoint from all earlier analyses, except that most \hp{} image pairs recur with new reference points and distractors. Repair test pairs are disjoint from all other item sets, and overlap between training, validation, and test splits was checked by pair identifier. Computing the final normalization and unembedding in fp32 removes exact ties between letter log-probabilities that occur in bf16. The answer format was selected on development data by requiring a median letter mass of at least $0.9$ and agreement of at least $95\%$ between greedy generation and the argmax letter. The items of Fig.~\ref{fig:qual} were selected before any map was computed: the item whose $\dref$ is closest to the task median and, within it, the distractor with the most attention on \refm{}. Every attention map reproduces the stored region measurements of \S\ref{sec:res_sel} and \S\ref{sec:mediation} to $1.4\times10^{-8}$.
\end{document}